\pdfoutput=1
\documentclass[11pt]{article}

\usepackage[preprint]{acl}
\usepackage{booktabs}
\usepackage[table]{xcolor}

\usepackage{times}
\usepackage{latexsym}
\usepackage[T1]{fontenc}
\usepackage[utf8]{inputenc}

\usepackage{microtype}
\usepackage{inconsolata}
\usepackage{graphicx}
\usepackage{booktabs}
\usepackage{array}
\usepackage{amsmath}
\usepackage{tikz}
\usepackage{placeins}
\definecolor{sanchez}{HTML}{E34948}
\definecolor{feijoo}{HTML}{2A78D6}
\definecolor{abascal}{HTML}{008300}

\newcommand{\rerun}[1]{#1}

\title{Sánchez, Feijóo, or Abascal? Speaker Attribution and Ideological Congruence in LLM Personas Reading Posts on X of the Leaders of the Three Main Parties of the 29N General Election in Spain}

\author{
  \textbf{Víctor Yeste\textsuperscript{1,2,*}},
  \textbf{Candida Maria Greco\textsuperscript{1,*}},
  \textbf{Paolo Rosso\textsuperscript{1,3}},
  \textbf{Andrea Tagarelli\textsuperscript{4}}
\\
\\
  \textsuperscript{1}PRHLT Research Center, Universitat Politècnica de València, Spain \\
  \textsuperscript{2}School of Science, Engineering and Design, Universidad Europea de Valencia, Spain \\
  \textsuperscript{3}Valencian Graduate School and Research Network of Artificial Intelligence (ValgrAI) \\
  \textsuperscript{4}DIMES Department, University of Calabria, Rende, Italy \\
  {\{vicyesmo, candida.greco\}@upv.es}, {prosso@dsic.upv.es}, {tagarelli@dimes.unical.it}
}

\begin{document}
\maketitle
{\renewcommand{\thefootnote}{*}\footnotetext{Equal contribution.}}

\begin{abstract}
Source cues are central to political communication, yet reception studies with large language model (LLM) personas as synthetic audiences have not tested whether personas respond to who speaks and not only to what is said. 
With Spain's general election of 29 November 2026 in mind, we run a speaker-attribution experiment using 1,075 anonymised X posts by the leaders of PSOE, PP and VOX on four topics (the economy and employment, household costs, immigration and corruption) and 791 distinct value-grounded, LLM-generated personas. %
Each post was assigned to 300 topic-selected persona profiles, 150 grounded in the Inglehart--Welzel cultural map and 150 in Moral Foundations Theory, under four conditions: no author, Sánchez, Feijóo or Abascal, yielding 322,470 complete persona--post quartets.
 Naming any leader lowered support for identical text, 
 with the penalty for Abascal nearly twice as large as those for Sánchez and Feijóo. The penalty follows the persona's elicited left--right self-placement (ideological congruence): left personas spare Sánchez and right personas spare Feijóo, although ideology was never  in the message-evaluation prompt. 
Corruption posts were penalised under every name, whereas on other topics the ideologically closest leader generally produced little or no penalty.
The leader order matches public opinion in 48 CIS barometers (2022--2026) in all three groups and the sign of all twelve differences, but fits barometers before the model's 2024 knowledge cutoff better than recent ones. 
Our findings demonstrate that LLM personas evaluate political speakers through systematic, value-congruent priors: 
however, while they replicate human candidate rank-orders across survey barometers, 
this alignment is ordinal rather than quantitative, 
highlighting their utility for controlled source-effect experiments though not as substitutes for empirical public opinion surveys. 
\end{abstract}

\section{Introduction}

In political communication the speaker is part of the message. The same proposal is judged differently when it comes from one's own party or from a rival, and party cues can outweigh the content of a policy \citep{cohen-2003-party, druckman-etal-2013-elite}. Large language models are now used to simulate political audiences, with personas standing in for groups of citizens. To our knowledge, \textbf{whether such personas respond to \emph{who} speaks, and not only to \emph{what} is said, has not been tested}, because reception studies with synthetic audiences have shown messages with their author either always visible or always hidden.  
The study we extend \citep{greco-etal-2026-morality} examined how value-grounded personas receive the posts of a single fixed anonymized actor. Because speaker identity did not vary, that design could not identify source-cue effects independently of message content.
This paper runs the classic source-cue experiment on synthetic audiences. We take 1,075 anonymised X posts by the leaders of Spain's three main parties, Pedro Sánchez (PSOE), Alberto Núñez Feijóo (PP) and Santiago Abascal (VOX), on four salient topics: the economy and employment, household costs, immigration and corruption. The experiment uses 791 distinct value-grounded personas instantiated with \texttt{gpt-oss-20b}; for each topic, 150 persona profiles grounded in the Inglehart--Welzel cultural map (IW; \citealp{inglehart-welzel-2005-modernization}) and 150 grounded in Moral Foundations Theory (MFT; \citealp{graham-etal-2013-moral}) evaluate every post under four conditions that differ only in the displayed author: no author, Sánchez, Feijóo or Abascal.
Because every post is shown under every name, the effect of a name is measured on identical text within the same audience. Spain is well suited to the question. Its party system has a centre-left government, a centre-right main opposition and a radical-right third party, so attribution can be crossed with ideological distance. The monthly barometers of the Spanish Centre for Sociological Research (Centro de Investigaciones Sociológicas, CIS) rate all three leaders and ask the left--right self-placement question we also put to the personas.

We ask four questions. \textbf{RQ1}: how does attributing the same message to Sánchez, Feijóo or Abascal, rather than presenting it anonymously, change its evaluation? \textbf{RQ2}: does the effect depend on the persona's elicited left--right self-placement, so that personas respond more positively to the leader closest to them? 
\textbf{RQ3}: Does the magnitude or direction of the attribution effect vary across salient political topics? \textbf{RQ4}: do left, centre and right personas rank the three leaders as the same groups of CIS respondents do?

Naming any leader lowered support for the same message, and Abascal's name did so about twice as much as the other two (\rerun{$-0.10$}, \rerun{$-0.10$} and \rerun{$-0.19$} on a $-1$ to $+1$ stance scale). The overall effect hides opposite reactions: left personas penalised Sánchez least and Abascal most, right personas penalised Feijóo least and Sánchez most, and all three congruence contrasts were positive, although the message-evaluation prompt never included the persona's elicited ideological placement. Only on corruption did every name cost support in every group. Elsewhere the closest leader's name carried no penalty. The order in which each group prefers the leaders is the order found in 48 CIS barometers from 2022 to 2026, in every group and in the sign of all twelve differences, but the match is dated: it holds in every barometer up to the model's June 2024 knowledge cutoff and weakens among right respondents from 2025.

Our contributions are: (1) the first fully crossed speaker-attribution design for synthetic audiences, with anonymisation and counterfactual attribution, complemented by a pre-experimental audit of whether the model recognises the leaders; 
(2) an elicited CIS-style self-placement whose associations with the personas' underlying moral and cultural values align directionally with relationships observed in human samples;
(3) a comparison with 48 CIS barometers, limited to order and direction, which also dates the personas' picture of the leaders; and (4) code, configuration, audit artifacts, post identifiers and derived labels, released without post text.\footnote{\url{https://github.com/VictorMYeste/spanish-political-communication}}

\section{Related Work}

\paragraph{LLM personas as synthetic audiences.}
Language models conditioned on demographic or biographical profiles have been used to simulate survey respondents \citep{argyle-etal-2023-out} and, after fine-tuning on survey data, to predict how subpopulations distribute their answers \citep{suh-etal-2025-language}. The limits of the approach are documented in the same literature: models reflect uneven defaults rather than the opinions of the groups they are asked to play \citep{santurkar-etal-2023-whose}, are harder to steer towards profiles whose traits rarely co-occur \citep{liu-etal-2024-evaluating-large}, flatten the groups they portray \citep{wang-etal-2025-large}, learn as well from irrelevant or misleading prompts as from instructive ones \citep{webson-pavlick-2022-prompt} and change what they simulate with the form of the persona cue \citep{lutz-etal-2025-prompt, weeber-etal-2026-one}. Synthetic samples can reproduce averages while missing the covariance structure of real opinion and adding a bias of their own \citep{bisbee-etal-2024-synthetic, boelaert-etal-2025-machine, von-der-heyde-etal-2026-vox}, and a review of 63 persona studies asks for personas grounded in empirical data and for a stated target population \citep{batzner-etal-2025-whose}. Two recent studies come close to our setting. Left- and right-leaning personas polarise the endorsement of persuasive content drawn from real media \citep{antelmi-etal-2026-political}, and political personas judge evidence by its fit with their assigned identity \citep{dash-etal-2026-persona}, but neither varies the speaker. Our own use of personas is narrower than survey simulation. Every comparison in this paper is between conditions judged by the same audience, and the comparison with survey data concerns order and direction only.

\paragraph{Source cues in political communication.}
That the messenger changes the reception of a message is one of the best-established results on political opinion. The same policy is judged differently when it carries a party label \citep{cohen-2003-party}, cues from the opposing party move evaluations even more than cues from one's own \citep{nicholson-2012-polarizing}, the effect grows with elite polarisation \citep{druckman-etal-2013-elite} and can override the respondent's stated ideology \citep{barber-pope-2019-party}, although citizens who do have policy information weigh it at least as much as the cue \citep{bullock-2011-elite}. Spain, which ranks among the most affectively polarised party systems in Europe \citep{reiljan-2020-fear, torcal-comellas-2022-affective}, is a strong case study. Exposure to the leaders' own accounts is a separate matter: following a candidate on Twitter did not increase affective polarisation in a panel quasi-experiment during the 2019 European Parliament campaign \citep{lorenzo-rodriguez-torcal-2022-twitter}. Language models already carry a view of these leaders: across 14 models, sentiment towards Sánchez is neutral to positive, towards Feijóo moderate and towards Abascal markedly negative \citep{goyanes-etal-2025-personality}. They are also sensitive to who is said to speak. Naming the source of a statement breaks the agreement between models that rate it \citep{germani-spitale-2025-source}, and a party cue changes the sentiment label that models give the same statement \citep{vallejo-vera-driggers-2025-llms}. A news outlet's name pulls the political bias that models assign to an unchanged article towards the outlet's known leaning \citep{park-kim-2025-measuring}. The politician named in a sentence shifts its sentiment classification, less so when the names are fictional \citep{elbouanani-etal-2025-analyzing}, and an audit of 1.9 billion data points in which only the entity name varies finds that models penalise right-wing politicians and favour left-wing ones \citep{elbouanani-etal-2026-scalable}. 
These studies generally frame source-induced shifts as model bias or annotation instability. We instead study them as persona-conditioned reception effects: holding message content fixed, we examine both the overall effect of a leader's name and how that effect varies across value-grounded persona profiles. This design distinguishes the common speaker-attribution effect from heterogeneity associated with persona values, although it does not eliminate the political priors of the underlying model \citep{feng-etal-2023-pretraining, rozado-2024-political, motoki-etal-2024-more, bernardelle-etal-2025-mapping}.

\paragraph{Value frameworks and moral reception.}
The personas are selected with two frameworks, Moral Foundations Theory with its MFQ-2 measure \citep{graham-etal-2013-moral, atari-etal-2023-morality} and the Inglehart--Welzel cultural map \citep{inglehart-welzel-2005-modernization}. Both are related to political ideology in human samples, where liberals and conservatives weigh the foundations differently \citep{graham-etal-2009-liberals, kivikangas-etal-2021-moral}, which lets us check whether the elicited placement of the personas follows their values. The persona resource is that of \citet{greco-lacava-tagarelli-2026-culturally}, who generate personas from World Values Survey variables and check their placement on the Inglehart--Welzel map and their moral profiles. The reception protocol comes from \citet{greco-etal-2026-morality}, who studied how such personas receive the messages of a single actor. This paper adds a speaker manipulation and an ideology layer to that protocol.

\section{Study Design}

Each selected persona $p$ evaluates every anonymised post $m$ in its assigned topic under four speaker conditions $c$: no author, Sánchez, Feijóo or Abascal. For each named condition $c \in \{\text{Sánchez},\text{Feijóo},\text{Abascal}\}$, the speaker-attribution effect is 
$\Delta_{p,m,c}=y_{p,m,c}-y_{p,m,\mathrm{anon}}$.
A persona--post pair enters the analysis only when all four conditions are observed, forming a complete quartet. Every effect is, therefore, a within-pair comparison.

The design is fully crossed: every post is shown under all three names, whoever wrote it. Therefore, differences between names cannot come from the posts, since each individual post appears in every condition, nor from the personas, since each persona judges the same posts in every condition, and must come from the name itself. The counterfactual attributions exist only inside the prompts: the real author of every post is known and used in the analysis, but never shown to the model.

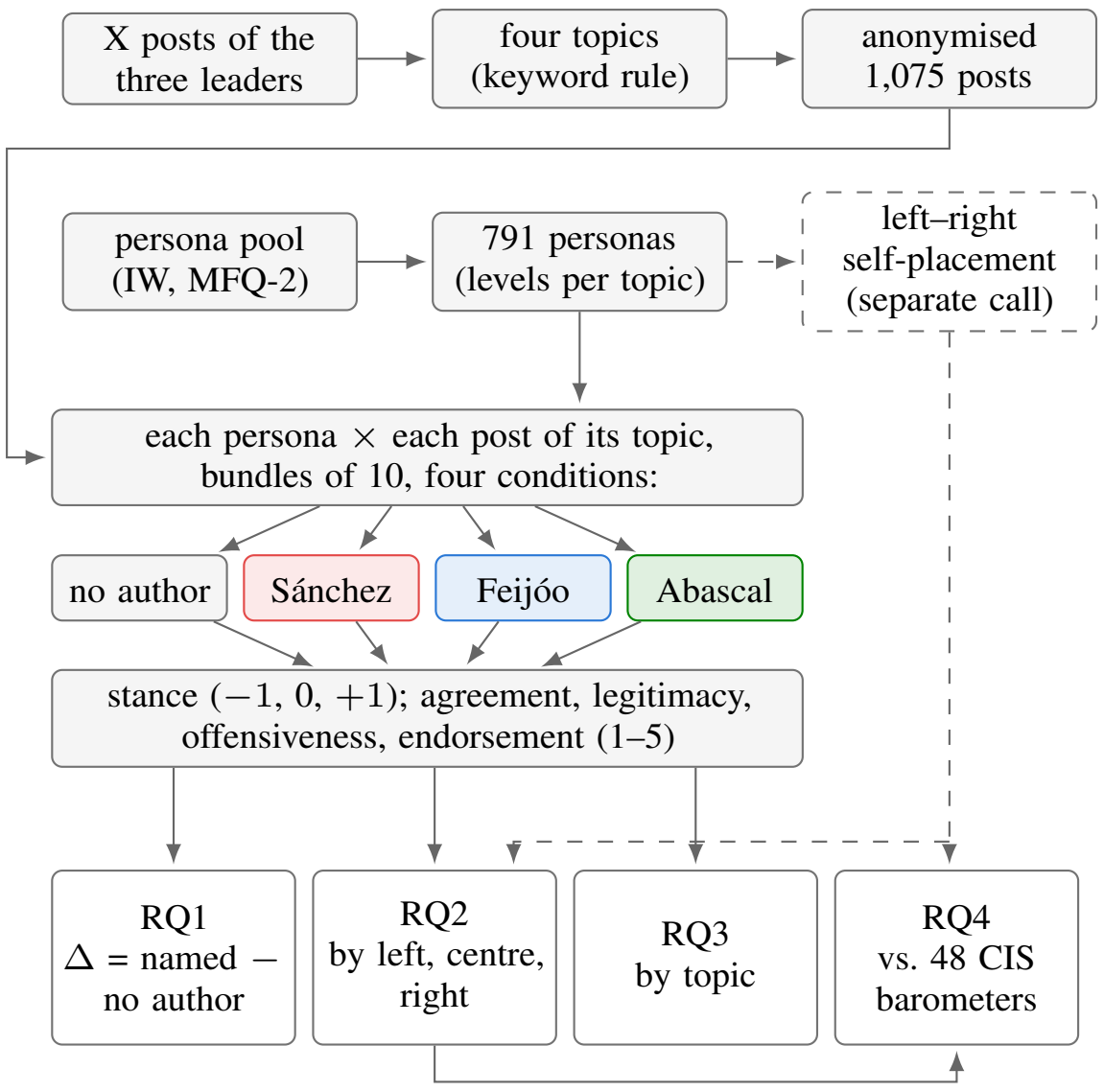
\begin{figure}[t]
\centering
\begin{tikzpicture}[
  font=\scriptsize,
  box/.style={draw=black!55, rounded corners=2pt, fill=black!4, align=center, inner sep=2.5pt, minimum height=6mm, text width=18.5mm},
  wide/.style={box, text width=50mm},
  rq/.style={box, text width=15mm, minimum height=12mm, fill=white},
  chip/.style={draw, rounded corners=2pt, align=center, inner xsep=1.5pt, inner ysep=2pt, minimum height=4.5mm, text width=11mm, text depth=0.3ex, text height=1.4ex},
  arr/.style={->, >=latex, draw=black!60, shorten >=1pt},
  darr/.style={->, >=latex, dashed, draw=black!60, shorten >=1pt}
]
\node[box] (posts) at (1.0,0) {X posts of the\\three leaders};
\node[box] (topics) at (3.55,0) {four topics\\(keyword rule)};
\node[box] (anon) at (6.1,0) {anonymised\\1,075 posts};
\draw[arr] (posts) -- (topics);
\draw[arr] (topics) -- (anon);
\node[box] (pool) at (1.0,-1.4) {persona pool\\(IW, MFQ-2)};
\node[box] (personas) at (3.55,-1.4) {791 personas\\(levels per topic)};
\node[box, dashed, fill=white] (ideo) at (6.1,-1.4) {left--right\\self-placement\\(separate call)};
\draw[arr] (pool) -- (personas);
\draw[darr] (personas) -- (ideo);
\node[wide] (judge) at (2.5,-2.75) {each persona $\times$ each post of its topic,\\bundles of 10, four conditions:};
\node[chip, draw=black!55, fill=black!4] (c0) at (0.515,-3.65) {no author};
\node[chip, draw=sanchez, fill=sanchez!12] (c1) at (1.838,-3.65) {Sánchez};
\node[chip, draw=feijoo, fill=feijoo!12] (c2) at (3.161,-3.65) {Feijóo};
\node[chip, draw=abascal, fill=abascal!12] (c3) at (4.484,-3.65) {Abascal};
\node[wide] (out) at (2.5,-4.55) {stance ($-1$, 0, $+1$); agreement, legitimacy,\\offensiveness, endorsement (1--5)};
\draw[arr] (anon.south) -- (anon.south |- 0,-0.62) -| (-0.4,-2.75) -- (judge.west);
\draw[arr] (personas.south) -- (judge.north -| personas.south);
\foreach \c in {c0,c1,c2,c3} {\draw[arr] (judge) -- (\c); \draw[arr] (\c) -- (out);}
\node[rq] (rq1) at (0.75,-6.2) {RQ1\\$\Delta$ = named $-$\\no author};
\node[rq] (rq2) at (2.55,-6.2) {RQ2\\by left, centre,\\right};
\node[rq] (rq3) at (4.35,-6.2) {RQ3\\by topic};
\node[rq] (rq4) at (6.15,-6.2) {RQ4\\vs.\ 48 CIS\\barometers};
\foreach \r in {rq1,rq2,rq3} {\draw[arr] (out.south -| \r.north) -- (\r.north);}
\draw[darr] (ideo.south) -- (rq4.north -| ideo.south);
\draw[darr] (ideo.south |- 0,-5.4) -| ([xshift=-3mm]rq2.north east);
\draw[arr] (rq2.south) -- ++(0,-0.25) -| (rq4.south);
\end{tikzpicture}
\caption{Study design. Anonymised posts by the three leaders are judged by value-grounded personas under four conditions that differ only in the displayed author. Each persona's left--right self-placement is elicited in a separate call (dashed) and never shown when it judges posts; it groups the personas for RQ2 and matches them to CIS respondents for RQ4.}
\label{fig:design}
\end{figure}

Our research questions are four cuts of $\Delta$ (Figure~\ref{fig:design}):  RQ1 averages it over all personas and posts for each name; RQ2 splits it by the persona's elicited left--right self-placement, obtained in a separate call that never enters the judgment prompt; RQ3 splits it by topic; and 
RQ4 compares the ranking of the three leaders induced by their mean speaker-attribution effects among left, centre and right personas with the ranking of their mean ratings among CIS respondents in the corresponding ideological groups.

\section{Data}
\subsection{Posts}

We collected every post published by the official X accounts of the three leaders between 1 April 2022, when Feijóo became leader of the PP, and 2 October 2026, through the full-archive search of the X API v2 and excluding retweets. This gave 10,863 posts, 7,625 of them original.\footnote{Eight posts counted by the API had been deleted before they could be downloaded. The corpus is the set of posts still public on 2 October 2026.} Only original posts are used as stimuli, because quotes and replies depend on another author's text, which would be missing from the stimulus or add a second, often identifiable, voice. We applied the quality filters of \citet{greco-etal-2026-morality}, which discard very short or low-information posts (here, under 40 characters or 7 words), and removed duplicates.

\subsection{Topics}

The topics were fixed by a rule before any data were downloaded, to tie them to public salience and to what the leaders actually post about. For each issue we averaged, over the 46 monthly CIS barometers in the window with a usable table, the share of respondents naming it among the three main problems of Spain (the CIS question asks for up to three), and we kept the most salient issues on which each leader had at least ten original posts. This selected four topics: (i) the economy and employment, (ii) household costs (housing, energy and prices), (iii) immigration and the border, and (iv) corruption. Health, third by salience, was dropped because Abascal had four posts on it.

Posts were matched to topics with a list of Spanish keywords and phrases for each topic, built for precision (Appendix~\ref{app:filters}), and posts matching no topic or two topics were left out. In a manual check of 80 matched posts, 96\% used the keyword in the topic's sense and 86\% were mainly about the topic. Table~\ref{tab:stimuli} gives the 1,075 stimuli by topic and real author. Two cells are thin (Abascal on household costs, Sánchez on corruption), but since every post is judged under all three names this limits only the analyses by real author.

\begin{table}[t]
\centering\footnotesize
\setlength{\tabcolsep}{3pt}
\begin{tabular}{lrrrr}
\toprule
Topic & Sánchez & Feijóo & Abascal & Total \\
\midrule
Economy, employment & 304 & 99 & 30 & 433 \\
Household costs & 123 & 94 & 11 & 228 \\
Immigration, border & 50 & 40 & 56 & 146 \\
Corruption & 27 & 127 & 114 & 268 \\
\midrule
Total & 504 & 360 & 211 & 1,075 \\
\bottomrule
\end{tabular}
\caption{Stimuli by topic and real author: the original X posts on the four topics that pass the quality and anonymisation checks. Every post is judged under all four conditions.}
\label{tab:stimuli}
\end{table}
\subsection{Anonymisation and attribution}

Before prompting, the three leaders, their parties, campaign hashtags and user handles were replaced by placeholders (\texttt{[POLITICAL\_ACTOR]}, \texttt{[PARTY]}, \texttt{[CAMPAIGN\_TAG]}, \texttt{[USER]}). Other politicians' names were kept, as replacing them would empty many posts, especially on corruption. Posts in which a name survived, or that became too short, were excluded, which left 7,299 of the 7,625 original posts (Appendix~\ref{app:filters}, Table~\ref{tab:appendix-filters}). All four conditions show the same anonymised text and differ only in the attribution line of the prompt.

Anonymisation hides names but not positions. A post that speaks for the Government fits Sánchez and no one else. A keyword rule for such markers (speaking for or against the Government, attacking the right, the left or the PP) flags 46\% of the stimuli as implausible under at least one name, though it misses more than it catches. Filtering on it would leave too few posts per cell, so we kept the fully crossed design and report the effect separately for true, plausible and implausible attributions (Appendix~\ref{app:markers}, Table~\ref{tab:appendix-plausibility}). %

\subsection{Personas}
\label{sec:personas}

We use the persona resource of \citet{greco-lacava-tagarelli-2026-culturally}, as \citet{greco-etal-2026-morality} did. Its personas are LLM-generated profiles built from ten World Values Survey variables, each with a short biography, Inglehart--Welzel (IW) coordinates and MFQ-2 moral-foundation scores \citep{inglehart-welzel-2005-modernization, atari-etal-2023-morality}. As in that paper, personas are selected under two frameworks, which also define what the model sees: IW personas are described by their biography and their survival--self-expression and traditional--secular-rational coordinates, and MFT personas by their biography, six foundation scores and 36 MFQ-2 item answers. 
For each topic and framework, we selected 50 personas at each of three topic-conditioned profile levels (\emph{high}, \emph{mid} and \emph{low}; Appendix~\ref{app:selection}). This gives 600 slots per framework and 791 framework-specific persona profiles (441 IW and 350 MFT), as some personas serve several topics.
 The levels are defined in the value space and never appear in the prompt (Appendix~\ref{app:selection}, Table~\ref{tab:appendix-profiles}). Ideology enters only through the self-placement described next.

The resource has no left--right variable, so we elicited one. Each persona answered the CIS self-placement question (from 1, furthest to the left, to 10, furthest to the right of Spanish politics) once, in a separate call with the same system prompt and persona text and no post. We grouped personas as left (1--4), centre (5--6) or right (7--10) and applied the same cut-offs to CIS respondents. The 1--10 scale has no single midpoint, so its two central values form the centre \citep{aybar-etal-2024-scale}, and studies of polarisation on the CIS series use the same left and right blocks \citep{torcal-carty-2026-polarizacion}. Figure~\ref{fig:placement} also reports each placement separately. The groups hold 27\%, 44\% and 28\% of IW personas and 39\%, 26\% and 35\% of MFT personas. The answer is the model's reading of the persona and serves only to group it, but it is far from arbitrary: although the persona descriptions contain no explicit ideological label and the elicited placement is never included in the message-evaluation prompt, self-placement follows the values shown to the model in directions consistent with relationships observed in human samples.
MFT personas sit further left with care and equality and further right with proportionality, authority and loyalty (Spearman $\rho$ from $-0.87$ to $0.67$), and IW personas sit further left with self-expression values ($\rho=-0.60$). The ideological variation used in RQ2 and RQ4 is 
therefore systematically associated with the value profiles shown to the model
(Appendix~\ref{app:values}).

\subsection{CIS barometers}

For RQ4 we use the microdata of the 48 monthly CIS barometers from April 2022 to October 2026, the period of the posts \citep{cis-2026-barometros}. Every barometer asks respondents to rate each leader from 1 to 10 and to place themselves on the same left--right scale as the personas. We kept the 175,764 of 193,263 respondents who placed themselves and rated all three leaders, computed weighted mean ratings within each barometer, and averaged the barometers with equal weight (Appendix~\ref{app:cis}).

\section{Experimental Setup}
\label{sec:setup}
\paragraph{Model and decoding.}
All judgments were generated with \texttt{gpt-oss-20b} \citep{openai-2025-gptoss}, an open-weight model with a stated knowledge cutoff of June 2024 and the model used by \citet{greco-etal-2026-morality}. It was served with vLLM through an OpenAI-compatible API, with temperature 0, a fixed seed, a limit of 4,096 tokens and JSON output. Each call presents a bundle of ten posts of its topic to one persona under one condition.
Generation was run on four NVIDIA A30 GPUs with 24 GB of memory each, with one independent vLLM replica serving the same model on each GPU. The economy-and-employment and immigration-and-border runs were completed on 27 October 2026, and the household-costs and corruption runs on 2 November 2026.

\paragraph{Prompt.}
The system prompt asks the model to role-play the persona, to treat the displayed attribution as given, and, when no speaker is named, not to infer one from style, topic or outside knowledge. The user prompt contains the persona profile, a line stating that the messages concern current political issues in Spain, the condition line (``every message below is attributed to [name]'' or ``no speaker identity is provided''), the evaluation rules, the ten anonymised messages and the output schema. Prompts are in English and the posts stay in Spanish. For each message the model returns a stance (support, neutral or reject) and four 1--5 ratings: agreement, legitimacy, offensiveness and endorsement (whether the persona would share the message as written), plus two short free-text fields that we do not analyse (full template in Appendix~\ref{app:prompt}, Figure~\ref{fig:appendix-prompt}). Of the 322,500 persona--post pairs, 322,470 (99.99\%) were observed in all four conditions and form the analysis set (Appendix~\ref{app:completeness}, Table~\ref{tab:appendix-completeness}).

\paragraph{Knowledge audit.}
Before the experiment, we checked that the model knows the three names. We asked it 20 times about each leader and about three control names, two invented but plausible Spanish names and one opaque code. It recognised each leader and named the correct party every time, placed Sánchez, Feijóo and Abascal at 3, 8 and 9 on the left--right scale with a spread of at most one point, and never recognised a control name (Appendix~\ref{app:audit}, Table~\ref{tab:appendix-audit}). This establishes the model's left-to-right ordering of the leaders used in RQ2 and shows that the attribution line carries the intended identities.

\paragraph{Estimation.}
Stance is coded $-1$, 0 and $+1$, and the ratings keep their 1--5 scale. We report means of $\Delta$ over complete quartets with 95\% confidence intervals two-way cluster-robust by message and persona \citep{cameron-etal-2011-robust}. Congruence is tested with the contrast $(\Delta_A-\Delta_B)_{\text{left}}-(\Delta_A-\Delta_B)_{\text{right}}$ for leaders A left of B, positive when the left group responds more favourably to the attribution of $A$ rather than $B$, relative to the right group. For the CIS we compute weighted mean ratings and leader differences within each barometer with linearised standard errors and pool the barometers with equal weight. Because a rating and a speaker-attribution effect are on different scales, we compare only the order of the leaders within each group and the sign of each difference.

\section{Results}

\subsection{Speaker attribution (RQ1)}

Naming any of the three leaders lowered support relative to the anonymous version of the same message (Table~\ref{tab:rq1-overall}). Sánchez and Feijóo produced almost identical effects ($\Delta=-0.100$ for both), whereas the penalty for Abascal was nearly twice as large ($\Delta=-0.191$); all three confidence intervals exclude zero. Most responses did not change category after attribution (approximately 65--70\%). Changes were nevertheless asymmetric: responses became less supportive more often than more supportive, particularly for Abascal (25.0\% versus 10.4\%). In absolute terms, half of the responses supported an anonymous post and a quarter rejected it (mean stance 0.24), so the penalties are not floor or ceiling effects. A name moved answers from support towards rejection while the neutral share stayed near a quarter: under Abascal's name, 40\% supported and 34\% rejected (Appendix~\ref{app:levels}, Table~\ref{tab:appendix-levels}).

The penalty extends beyond stance: in all 24 framework--topic--speaker cells, naming the leader lowered agreement, legitimacy and endorsement and raised perceived offensiveness (Appendix~\ref{app:rq1}, Table~\ref{tab:appendix-secondary}). It is smallest when the displayed name is the true author (\rerun{$-0.07$} to \rerun{$-0.09$} by framework) and largest for implausible counterfactual attributions (\rerun{$-0.15$} and \rerun{$-0.23$}), though this comparison is descriptive, since the three categories hold different posts (Appendix~\ref{app:markers}, Table~\ref{tab:appendix-plausibility}).

\begin{table}[t]
\centering\footnotesize
\setlength{\tabcolsep}{3.5pt}
\begin{tabular}{lcrr}
\toprule
Speaker & $\Delta$ stance [95\% CI] & Less supp. & Same \\
\midrule
Sánchez & -0.100 [-0.114, -0.086] & 18.6\% & 69.6\% \\
Feijóo & -0.100 [-0.112, -0.087] & 18.9\% & 69.5\% \\
Abascal & -0.191 [-0.206, -0.175] & 25.0\% & 64.6\% \\
\bottomrule
\end{tabular}
\caption{Speaker-attribution effects on stance. $\Delta$ is the mean attributed-minus-anonymous difference over complete quartets, both frameworks and all topics; 95\% confidence intervals two-way cluster-robust by message and persona. The last two columns give the share of responses that became less supportive or stayed the same.}
\label{tab:rq1-overall}
\end{table}

\begin{figure*}[t]
\centering
\includegraphics[width=\textwidth]{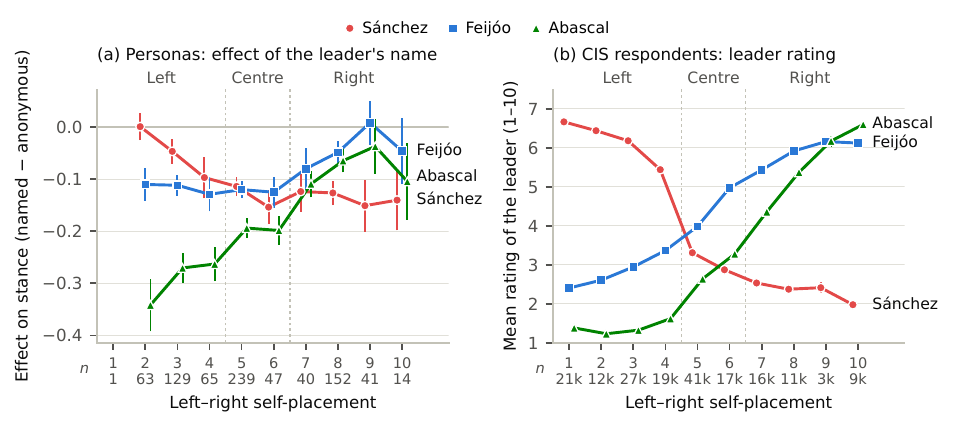}
\caption{Leader preferences by left--right self-placement. (a) Personas: speaker-attribution effect on stance at each elicited placement, both frameworks and all topics, with 95\% confidence intervals; placement 1 (one persona) is not shown. (b) CIS respondents: mean rating of each leader (1--10) at each self-placement, pooled over the 48 barometers from April 2022 to October 2026, with 95\% confidence intervals. $n$ is the number of personas (a) and respondents (b) at each placement. Dotted lines separate the left (1--4), centre (5--6) and right (7--10) groups.}
\label{fig:placement}
\end{figure*}

\begin{table*}[t]
\centering\small
\setlength{\tabcolsep}{4pt}
\begin{tabular}{lcccc}
\toprule
Leaders A -- B & Left & Centre & Right & Left $-$ right \\
\midrule
Sánchez -- Feijóo & 0.069 [0.055, 0.084] & -0.002 [-0.013, 0.009] & -0.080 [-0.097, -0.064] & 0.150 [0.128, 0.172] \\
Sánchez -- Abascal & 0.241 [0.212, 0.271] & 0.072 [0.053, 0.090] & -0.057 [-0.079, -0.036] & 0.298 [0.264, 0.333] \\
Feijóo -- Abascal & 0.172 [0.152, 0.192] & 0.074 [0.059, 0.089] & 0.023 [0.007, 0.040] & 0.148 [0.124, 0.173] \\
\bottomrule
\end{tabular}
\caption{Congruence contrasts on stance. The first three columns report $\Delta_A-\Delta_B$ within each group; positive values mean that the group penalised A less than B. The last column is the difference between the left and right groups, positive under congruence. Cells report the estimate [95\% CI].}
\label{tab:rq2-congruence}
\end{table*}

\subsection{Ideological congruence (RQ2)}

Figure~\ref{fig:placement}a shows the effect of each name at every elicited placement, and Table~\ref{tab:rq2-congruence} the congruence contrasts. Group estimates are in Appendix~\ref{app:rq2}, Table~\ref{tab:appendix-rq2-groups}.

The near-identical overall effects of Sánchez and Feijóo in RQ1 (Table~\ref{tab:rq1-overall}) are the average of opposite reactions. Left personas penalised Sánchez least ($\Delta=-0.047$), Feijóo more ($-0.116$) and Abascal most ($-0.288$). Right personas reversed the two main leaders, penalising Feijóo least ($-0.049$), then Abascal ($-0.072$) and Sánchez most ($-0.129$), while centre personas penalised Sánchez and Feijóo equally ($-0.123$ and $-0.121$) and Abascal more ($-0.195$). The change across the scale is gradual (Appendix~\ref{app:rq2}, Table~\ref{tab:appendix-rq2-placement}). From placement 2 to 9, Sánchez's effect goes from $0.001$ to $-0.151$, Feijóo's from $-0.110$ to $0.008$ and Abascal's from $-0.342$ to $-0.037$, and Sánchez's line crosses Feijóo's between 5 and 6 and Abascal's between 6 and 7, so the result does not hinge on the cut-offs.

The congruence contrasts order the leaders as the model placed them in the knowledge audit (Sánchez, Feijóo, Abascal), and all three are positive with confidence intervals above zero (Table~\ref{tab:rq2-congruence}), the largest for the two leaders furthest apart, Sánchez and Abascal (0.298). Even right personas penalised Abascal slightly more than Feijóo (0.023), consistent with most of them placing themselves at 7 or 8, closer to Feijóo. The pattern holds under both frameworks, in all four topics and on all five outcomes (Appendix~\ref{app:rq2}, Tables~\ref{tab:appendix-rq2-congruence-topic} and~\ref{tab:appendix-rq2-outcomes}). Only two of the 24 framework--topic contrasts have intervals that include zero, both among IW personas (Sánchez--Feijóo on corruption, Feijóo--Abascal on immigration).

Congruence also qualifies the general penalty of RQ1. Outside corruption, attribution to the ideologically closest leader produced no detectable penalty: for left personas, Sánchez's name increased support for economy messages ($\Delta=0.040$ [0.026, 0.055]) and did not significantly change support for household-cost or immigration messages; for right personas, Feijóo's name did not significantly change support on any of the three topics. Corruption is the exception. There every group penalised every leader, and left personas penalised Sánchez ($-0.338$) as much as Feijóo ($-0.329$).

Since the elicited ideological placement was never included in the message-evaluation prompt and was systematically associated with the personas' values (Section~\ref{sec:personas}), the congruence pattern is associated with the value-grounded profiles rather than with an explicit ideological label.

\begin{table*}[t]
\centering\small
\setlength{\tabcolsep}{5pt}
\begin{tabular}{lrrrrrr}
\toprule
& \multicolumn{3}{c}{IW personas} & \multicolumn{3}{c}{MFT personas} \\
\cmidrule(lr){2-4}\cmidrule(lr){5-7}
Topic & Sánchez & Feijóo & Abascal & Sánchez & Feijóo & Abascal \\
\midrule
Economy and employment & -0.030 & -0.055 & -0.186 & -0.008 & -0.042 & -0.182 \\
Household costs & -0.051 & -0.057 & -0.168 & -0.053 & -0.047 & -0.132 \\
Immigration and border & -0.056 & -0.030 & -0.164 & -0.065 & -0.031 & -0.101 \\
Corruption & -0.194 & -0.173 & -0.202 & -0.395 & -0.348 & -0.335 \\
\bottomrule
\end{tabular}
\caption{Speaker-attribution effects on stance by topic and persona-selection framework. Negative values mean that the name reduced support relative to the anonymous version of the same message. Confidence intervals in Appendix~\ref{app:rq1}, Table~\ref{tab:appendix-topic-ci}.}
\label{tab:rq3-topics}
\end{table*}

\begin{table*}[t]
\centering\small
\setlength{\tabcolsep}{4pt}
\begin{tabular}{lrccccccc}
\toprule
& & \multicolumn{3}{c}{CIS mean rating} & & & Barometers with \\
\cmidrule(lr){3-5}
Group & Respondents & Sánchez & Feijóo & Abascal & CIS order & Persona order & the same order \\
\midrule
Left (1--4) & 78,751 & 6.23 & 2.83 & 1.40 & S $>$ F $>$ A & S $>$ F $>$ A & 48/48 \\
Centre (5--6) & 57,904 & 3.19 & 4.26 & 2.82 & F $>$ S $>$ A & F $>$ S $>$ A & 39/48 \\
Right (7--10) & 39,109 & 2.33 & 5.82 & 5.43 & F $>$ A $>$ S & F $>$ A $>$ S & 32/48 \\
\bottomrule
\end{tabular}
\caption{Order of the three leaders by ideological group among CIS respondents and personas. CIS: weighted mean rating (1--10) over the 48 barometers from April 2022 to October 2026, each barometer weighing the same; all 95\% confidence intervals are within $\pm0.04$. Personas: order of the speaker-attribution effects on stance (Figure~\ref{fig:placement}a). S: Sánchez; F: Feijóo; A: Abascal. The last column counts the barometers whose order is the same as the personas'.}
\label{tab:rq4-orders}
\end{table*}

\subsection{Topics (RQ3)}
The magnitude of the attribution penalty varies across topics
(Table~\ref{tab:rq3-topics}, with intervals in Appendix~\ref{app:rq1}, Table~\ref{tab:appendix-topic-ci}). Corruption is the costliest topic for every name: attribution effects ranged from $-0.20$ to $-0.17$ among IW personas and from $-0.40$ to $-0.34$ among MFT personas. In absolute terms, corruption posts were close to neutral without an author ($-0.04$) and a majority rejected them under any name (54--57\%), whereas household-cost posts kept majority support under every name (Appendix~\ref{app:levels}, Table~\ref{tab:appendix-levels}). On the economy, household costs and immigration, the effects of Sánchez's and Feijóo's names were comparatively small ($-0.07$ to $-0.01$), while Abascal's name produced penalties between $-0.19$ and $-0.10$ on each of those three topics, so his penalty is not specific to corruption. Only one of the 24 framework--topic--speaker estimates had a confidence interval that included zero: Sánchez on economy and employment among MFT personas ($\Delta=-0.008$, 95\% CI $[-0.024, 0.008]$).

The aggregate topic pattern reflects the group-specific reactions observed in RQ2
(Appendix~\ref{app:rq2}, Table~\ref{tab:appendix-rq2-topic-groups}). Outside corruption, the small overall effects of Sánchez and Feijóo
reflect a penalty from the opposite ideological group and little or no detectable penalty from their own side.
Abascal's penalty comes mostly from left and centre personas, who penalise his name by $-0.14$ to $-0.29$ on those topics, whereas right personas leave it near zero on household costs ($-0.002$). On corruption every group penalised every leader, including its own side. The attribution effect therefore depends on what the message says as well as on who is said to say it.

\subsection{Comparison with CIS respondents (RQ4)}

Over the whole period, the order of the leaders is the same among personas and CIS respondents in all three groups (Table~\ref{tab:rq4-orders}), and all nine differences between two leaders within a group, as well as the three left-minus-right contrasts, have the same sign among personas and respondents (Appendix~\ref{app:cis}, Table~\ref{tab:appendix-rq4-pairs}). The weakest agreement is in the centre, where respondents rate Feijóo 1.06 points above Sánchez while centre personas barely separate the two (0.002, interval including zero).

The agreement is not uniform across the scale (Figure~\ref{fig:placement}b against~\ref{fig:placement}a, and Appendix~\ref{app:cis}, Table~\ref{tab:appendix-rq4-placement}). Personas and respondents rank the leaders in the same order at placements 1--4, 7 and 8. Where they differ, the personas are less favourable to Abascal or more favourable to Sánchez than respondents at the same placement, most clearly at 9 and 10, where respondents rate Abascal as high as Feijóo or higher while personas still prefer Feijóo, though only 41 and 14 personas sit there.

The agreement also depends on the period. Among right respondents the order matches the personas' in every barometer from April 2022 to December 2024, but in only 4 of the 20 since January 2025. Their Feijóo-minus-Abascal difference fell from about one point in 2022--2023 to below zero in 16 of those 20 barometers (Appendix~\ref{app:cis}, Figure~\ref{fig:appendix-rq4-trend}). Centre respondents show a weaker version of the same change, matching every barometer to December 2024 and 11 of the 20 since. Therefore, the personas resemble Spanish opinion of 2022--2024 more closely than that of the latest barometers. The timing is consistent with the model's June 2024 knowledge cutoff (Section~\ref{sec:setup}). In all 23 barometers up to that month, the CIS order matches the personas' in all three groups. The comparison cannot show that the cutoff is the reason, and, like any external benchmark, it shows only that the synthetic pattern resembles the survey pattern in order and direction. It says nothing about whether the personas represent Spanish citizens.

\section{Discussion and Conclusion}

The name changes the judgment of identical text, and in the direction that source-cue research on human respondents would predict. 
Outside corruption, attribution to the ideologically closest leader produced little or no penalty, whereas attribution to a rival reduced support. Overall, Abascal's name produced the largest penalty, consistent with the negative sentiment towards Abascal expressed by language models \citep{goyanes-etal-2025-personality} and with the broader penalty for right-wing politicians documented by \citet{elbouanani-etal-2026-scalable}. The model also received Abascal's own posts worst when they carried no name. Read anonymously, they had a mean stance of $-0.35$, against 0.47 for Sánchez's posts and 0.28 for Feijóo's, and they were the lowest in every ideological group and under both persona-selection frameworks (Appendix~\ref{app:levels}, Table~\ref{tab:appendix-levels-author}). This difference in content is larger than any attribution effect, but it does not enter $\Delta$, because every post is compared with itself under each name.
The pattern is one of repulsion more than attraction. Personas were rarely more favourable to a message because their own side signed it, but they were less favourable when the other side did, as human respondents react more to out-party than to in-party cues \citep{nicholson-2012-polarizing}. That any name lowers support relative to the anonymous version suggests that an attributed statement may be read as a partisan act. The reading is harshest on corruption, where every group penalised every leader, including its own.

\textbf{The personas were not assigned an explicit political ideology:
they were conditioned on biographies and value profiles}, while their
left--right placement was elicited afterwards and was never shown
during message evaluation. That placement was systematically associated with the MFQ-2 and Inglehart--Welzel dimensions in directions consistent with relationships observed in human samples.
\citet{greco-etal-2026-morality} found that value-grounded personas differ in how they receive the same messages. Here that differentiation turns out to be politically organised, so that a value profile carries a position on who may say what.

The comparison with the CIS shows how far that picture resembles Spanish opinion, and where it stops. Order and direction match in every group and every pairwise difference, and scale cannot be compared at all, since a speaker-attribution effect on a message and a rating of a leader are different quantities. The match is also dated. The personas reproduce the ordering of the barometers up to the model's knowledge cutoff and miss the movement of right respondents towards Abascal since 2025. 
A synthetic audience may carry dated political associations inherited from its training data.
Before personas are used as an audience in a live campaign, the date of the picture they carry should be checked, as RQ4 does here, and their output should not be read as current opinion.

  Although the pooled leader order matches in all three ideological groups, agreement is not perfect wave by wave: the persona order matches 48 of 48 barometers for the left, 39 of 48 for the centre, and 32 of 48 for the right. Moreover, ordering only three ideologically separated leaders is a low-dimensional benchmark. The match may reflect broad learned associations about the Spanish political spectrum rather than fine-grained simulation of voter behaviour, and should therefore be treated as a limited external-validity check rather than evidence of survey equivalence. 

Value-grounded personas respond to who speaks as well as to what is said, coherently and along ideological lines. That makes them useful for studying source effects, and it does not make them stand-ins for voters. Several extensions follow. The most immediate is to repeat the experiment with two further open-weight models, to see whether the pattern is specific to \texttt{gpt-oss-20b}. Labelling the posts by electoral period would show whether the speaker effect grows during campaigns, and asking the personas the CIS leader-rating item would make a like-with-like comparison of scale possible. The keyword lower bound on how much an anonymised post still reveals its author could be replaced by a model-based plausibility check of each attribution.

\section*{Limitations}

The results come from one model, \texttt{gpt-oss-20b}, whose knowledge ends in June 2024 while the posts run to October 2026. The size of the penalties, the concentration of IW personas at the scale midpoint and the picture of the leaders may all be specific to it. 
Posts about later events are judged by a model that does not know them, and its representation of the three leaders reflects information available only up to mid-2024, as the divergence from right-wing CIS respondents after 2025 suggests.
Two further open-weight models are planned.

The personas are generated profiles, not people, and their ideology is one elicited answer each. We did not test the stability of that answer under rewordings of the question, and a scale anchored to Spanish politics was put to personas most of whom are not Spanish. Among IW personas 41\% answered exactly 5, against 22\% of CIS respondents (Appendix~\ref{app:ideology}), so the centre group is heterogeneous, although the by-placement results do not depend on the cut-offs.

We evaluate value-grounded personas based on IW and MFT profiles,  
but do not benchmark them against alternative steering methods like simple demographic prompting or zero-shot baselines. 
Because persona behavior in LLMs is highly prompt-dependent, our results characterize value-grounded synthetic audiences specifically, without extending to other persona conditioning techniques.

The anonymous baseline is not fully anonymous. Names and party labels were removed, but the content of a post often reveals its author's position. Abascal's posts, for instance, were received far worse than the others' even without a name (Appendix~\ref{app:levels}), which may partly reflect recognition of their author. The keyword flags that mark 46\% of the stimuli as implausible under some name are a lower bound, and we ran no model-based plausibility or leakage probe. The design stays fully crossed, so the counterfactual attributions include implausible ones, and the split by true, plausible and implausible attribution is descriptive, because the three categories hold different posts.

  The fully crossed design also creates source–content combinations that are implausible and, in some cases, internally inconsistent
The especially large penalties in these cases cannot be attributed uniquely to partisan rejection: they may also reflect sensitivity to source–content mismatch or difficulty sustaining the assigned role. Future work should separate these mechanisms using independent plausibility ratings and explicit contradiction measures. 

The stimuli are original X posts on four topics (the economy and employment, household costs, immigration and corruption), assigned by a keyword rule whose main-topic precision was 86\% in a manual check, with thin cells for Abascal on household costs and Sánchez on corruption. The design has one leader on the left and two on the right. The space to the left of the PSOE had no single leader over the period: Yolanda Díaz stepped down as leader of Sumar in June 2024, and the parties of the Frente Amplio named Ada Colau their candidate on 12 October 2026, after the posts were collected. That space is the natural place for a fourth leader, and the CIS rates Díaz in all 48 barometers. The prompt template and the persona profiles are in English while the posts are in Spanish, and a fully Spanish run is left for future work.

The CIS comparison is limited to order and sign, since a speaker-attribution effect on a message and a rating of a leader cannot be put on one scale. The CIS intervals ignore the clustering of the sample design, and agreement in order does not make the personas representative of Spanish citizens.

\section*{Ethical considerations}

The study uses public posts by three public officials, collected through the official X API under its terms, and the public microdata of the CIS, reused under its conditions of use with attribution. Public statements by politicians are still personal data under the GDPR. They were processed for academic research only, stored in institutional storage, and not used to profile any private individual. We release code, configuration, audit artifacts, post identifiers and derived labels, but no post text.

Anonymisation is an experimental control. We make no claim that the posts are private or unidentifiable, and the counterfactual attributions exist only inside the prompts: the paper never presents a post as a statement by a leader who did not write it, and the example in Appendix~\ref{app:anonymisation}, Table~\ref{tab:appendix-example} names its real author.

The main ethical risk is overinterpretation. The personas are not voters, their elicited placement is the model's reading of a value profile, the left, centre and right groups make no claim about real electorates, and the resemblance to the CIS concerns the order of the leaders only. Since this preprint appears before a general election, we stress that nothing in it forecasts votes, evaluates the leaders or measures public support for them. The attribution penalties describe the behaviour of one model and say nothing about what the public thinks.

\section*{Acknowledgments}

This work was partially supported by the following projects: ELENI: Estrategia Local En Narrativas Inteligente, funded by CDTI (IDI-20260067); Multimodal and Multisensor data-centric AI models (ANNOTATE-MULTI2), grant PID2024-156022OB-C32 funded by MICIU/AEI/10.13039/501100011033 and by ERDF/EU. %

The authors used AI assistants (Claude, Anthropic) for language editing, for writing and checking the analysis and figure scripts, and for an informal reading of samples of role-marker matches (Appendix~\ref{app:markers}). The design, the analyses and the claims are the authors' own, and all text was reviewed by them.

\bibliography{custom}

\appendix
\setcounter{dbltopnumber}{4}\renewcommand{\dbltopfraction}{0.9}
\setcounter{topnumber}{4}\renewcommand{\topfraction}{0.9}
\setcounter{totalnumber}{6}\renewcommand{\textfraction}{0.1}
\renewcommand{\dblfloatpagefraction}{0.7}\renewcommand{\floatpagefraction}{0.7}

\section{Stimuli, Prompts and Attribution Checks}
\label{app:stimuli}

\subsection{Corpus and filters}
\label{app:filters}

Table~\ref{tab:appendix-filters} gives the quality and anonymisation checks by leader. The text is cleaned first (HTML entities decoded, URLs removed, whitespace collapsed), and the thresholds are those of the previous paper: fewer than 40 characters or fewer than 7 words. Its list of exact low-information texts is not needed, because such texts are already below 7 words, and its truncation flag does not apply, because the X API returns the full text. After anonymisation, two checks follow: an identity leak (a leader's or party's name still present, for instance through a misspelling or a missing space) and degradation (fewer than 40 characters, fewer than 7 words not counting placeholders, or more than 3 placeholders in 80 characters or less). A duplicate is a post whose anonymised text, ignoring case, repeats an earlier one that passed every check. A post can fail several checks; most degraded posts were already too short, and only 24 fail because of degradation alone (event announcements such as ``Este viernes 28 de abril en \#Badajoz [CAMPAIGN\_TAG]''). The excluded posts are mostly greetings, slogans and image-only posts, and only 8 of the 326 have one of the four topics.

\begin{table}[htbp]
\centering\footnotesize
\setlength{\tabcolsep}{3pt}
\begin{tabular}{lrrrr}
\toprule
& Sánchez & Feijóo & Abascal & All \\
\midrule
Original posts & 3,295 & 3,032 & 1,298 & 7,625 \\
\quad Under 40 characters & 31 & 90 & 90 & 211 \\
\quad Under 7 words & 31 & 90 & 137 & 258 \\
\quad Identity leak & 0 & 0 & 2 & 2 \\
\quad Degraded text & 36 & 109 & 154 & 299 \\
\quad Duplicate text & 7 & 6 & 10 & 23 \\
Pass every check & 3,252 & 2,915 & 1,132 & 7,299 \\
\bottomrule
\end{tabular}
\caption{Quality and anonymisation checks on the original posts. A post can fail several checks.}
\label{tab:appendix-filters}
\end{table}

Table~\ref{tab:appendix-topics} gives the topic rule and its outcome. Salience is the mean share of respondents who named the issue among the three main problems of Spain, over the 46 barometers of the window with a usable table (no barometer was published in August, nor in June and September 2023, around the general election, and the May 2023 and December 2024 tables could not be parsed). Before ranking we excluded the categories that name politics itself as the problem (the government and parties, the behaviour of politicians, the lack of agreement), since they point to no policy area and overlap with the evaluation of politicians that the experiment measures. Diffuse categories (the crisis of values, uncertainty about the future) and non-substantive answers were excluded as well. The economy, unemployment and job quality were merged into one topic before the data were downloaded, as were housing, energy and prices; the ranking is the same when the merges are undone one at a time, except that splitting the economy into its three categories pushes corruption to seventh. The sixth issue, education, had a mean salience of 7.1\%.

Posts were assigned to the five issues with a Spanish keyword registry, matched after removing URLs and accents, with a few phrases removed first because they contain a topic word with another meaning (``cordón sanitario'', ``inquilino de la Moncloa'', homes evacuated in fires and floods). In the spot check of 80 matched posts (20 per retained topic, one annotator), the matched term was used in the topic's sense in 19, 18, 20 and 20 of 20 posts on the economy, household costs, immigration and corruption, and the post was mainly about the topic in 17, 16, 19 and 17. The three word-sense errors were fixed in the registry; the remaining errors are passing mentions that keywords cannot separate (``empleo'' or ``vivienda'' in a list of laws, ``corrupción'' in posts about the floods or the Gaza war).

\begin{table}[htbp]
\centering\footnotesize
\setlength{\tabcolsep}{3pt}
\begin{tabular}{lrrrrr}
\toprule
Topic & CIS \% & Sánchez & Feijóo & Abascal & Kept \\
\midrule
Economy & 63.2 & 305 & 100 & 30 & yes \\
Household & 22.9 & 123 & 95 & 11 & yes \\
Health & 14.6 & 92 & 56 & 4 & no \\
Immigration & 13.8 & 50 & 40 & 57 & yes \\
Corruption & 8.3 & 27 & 128 & 117 & yes \\
\midrule
Several topics & & 55 & 33 & 32 & \\
No topic & & 2,643 & 2,580 & 1,047 & \\
\bottomrule
\end{tabular}
\caption{Topic rule: the five most salient issues in the CIS barometers (CIS \%: mean share of respondents naming the issue, April 2022 to October 2026) and the original posts matching each one, before the quality checks. An issue is kept when every leader has at least ten posts; health fails on Abascal. Posts matching two issues or none were not used as stimuli.}
\label{tab:appendix-topics}
\end{table}

\subsection{Anonymisation}
\label{app:anonymisation}

Table~\ref{tab:appendix-placeholders} lists what each placeholder replaces. Other politicians' names stay (Ábalos, Ayuso, Puigdemont, Trump): replacing them would empty many posts, especially on corruption, and the role-marker check of Appendix~\ref{app:markers} covers the cases in which they give the author away. Bare first names stay too, because ``Pedro'' is mostly someone else, as do hashtags about content, places and dates (\#DANA, \#COP28, \#23J) and party adjectives (``socialistas''). About half of the posts that pass every check contain a placeholder (Sánchez 1,348 of 3,252, Feijóo 1,516 of 2,915, Abascal 747 of 1,132); \texttt{[POLITICAL\_ACTOR]} appears in 157, 661 and 283 of them, and in the posts of Feijóo and Abascal it almost always stands for Sánchez. Table~\ref{tab:appendix-example} shows one post as the model saw it.

\begin{table}[htbp]
\centering\footnotesize
\setlength{\tabcolsep}{3pt}
\begin{tabular}{>{\ttfamily}l>{\raggedright\arraybackslash}p{4.6cm}}
\toprule
\normalfont Placeholder & Replaces \\
\midrule
{[POLITICAL\_ACTOR]} & The three leaders: full names, surnames, ``Pérez-Castejón'', ``sanchismo'', ``sanchista'' and their own accounts \\
{[PARTY]} & PSOE, PP, VOX and their regional branches (PSC, PSdeG, PSPV, \ldots), ``Partido Popular'', ``Partido Socialista'', PPE; party accounts and European party families \\
{[CAMPAIGN\_TAG]} & Hashtags with a leader's or party's name (\#VotaPSOE, \#InvestiduraFeijóo, \#SánchezDimisión) and 54 party slogans chosen by hand from the 203 hashtags used at least three times (\#EsElMomento, \#ElGobiernoDeLaGente, \#SentidoComún) \\
{[USER]} & Every other account, except 27 public institutions that keep a plain name (@Congreso\_Es becomes ``Congreso de los Diputados'', @NATO becomes ``OTAN'') \\
\bottomrule
\end{tabular}
\caption{Anonymisation placeholders.}
\label{tab:appendix-placeholders}
\end{table}

\begin{table}[htbp]
\centering\footnotesize
\begin{tabular}{>{\raggedright\arraybackslash}p{0.94\columnwidth}}
\toprule
\textbf{Original} (Feijóo, 31 January 2023; the image link at the end is removed by the cleaning step) \\
Pedro Sánchez presume de creación de empleo y de crecimiento económico. La única realidad es que España está a la cola de la Unión Europea. \\
\midrule
\textbf{Anonymised} (shown in all four conditions) \\
\texttt{[POLITICAL\_ACTOR]} presume de creación de empleo y de crecimiento económico. La única realidad es que España está a la cola de la Unión Europea. \\
\midrule
\textbf{Condition line, no author} \\
No speaker identity is provided in this condition. Evaluate every message without assuming who wrote it. \\
\addlinespace
\textbf{Condition line, attributed} (Pedro Sánchez, Alberto Núñez Feijóo or Santiago Abascal) \\
For this experimental condition, every message below is attributed to Pedro Sánchez. Treat this attribution as given while evaluating the message. \\
\bottomrule
\end{tabular}
\caption{One stimulus and the four condition lines. The role-marker rule flags this post as implausible under Sánchez's name, since it attacks him.}
\label{tab:appendix-example}
\end{table}

\subsection{Prompt template}
\label{app:prompt}

Figure~\ref{fig:appendix-prompt} gives the system prompt and the user prompt of a call. Each call presents ten anonymised posts of one topic to one persona under one condition; the four conditions differ only in the condition line. The persona profile is the biography and the values of the persona's framework, in the format of the previous paper. The ideology call of Appendix~\ref{app:ideology} uses the same system prompt and persona profile, followed by the self-placement question and a one-field schema.

\begin{figure*}[tbp]
\small
\begin{verbatim}
SYSTEM
Role-play the person described in the supplied persona profile. Evaluate Spanish political
messages as that person would, using the displayed speaker attribution as part of the
experimental context. Treat a displayed attribution as given and do not question whether it
is historically correct. When no speaker is named, do not infer one from style, topic, or
outside knowledge. Do not try to recover or mention the hidden real author. Separate the
substantive position from rhetorical tone. Return only valid JSON with no markdown.

USER
[PERSONA PROFILE]
Survey respondent profile.
Biography: <biography>
  IW personas:  Cultural positioning in the Inglehart-Welzel space: surv-self = <x>,
                trad-sec = <y>. Interpretation hint: lower surv-self is closer to
                survival/material-security orientation; higher surv-self is closer to
                self-expression orientation. Lower trad-sec is closer to traditional
                values; higher trad-sec is closer to secular-rational values.
  MFT personas: [MORAL PROFILE] Moral profile from MFQ-2 on a 1..5 scale. Higher values
                indicate that the statement describes the person more strongly.
                Foundation-level MFQ-2 scores: <six foundation scores>. Direct MFQ-2 item
                responses are provided below. Use them as concrete evidence of the
                person's moral intuitions. MFQ-2 response scale: Does not describe me at
                all (1); Slightly describes me (2); Moderately describes me (3); Describes
                me fairly well (4); Describes me extremely well (5).
                MFQ-2 item responses:
                1. <item 1> Response: <1-5>  ...  36. <item 36> Response: <1-5>

[POLITICAL CONTEXT]
The messages concern current political issues in Spain.
  No author:  No speaker identity is provided in this condition. Evaluate every message
              without assuming who wrote it.
  Attributed: For this experimental condition, every message below is attributed to
              <Pedro Sánchez | Alberto Núñez Feijóo | Santiago Abascal>. Treat this
              attribution as given while evaluating the message.

[EVALUATION RULES]
- stance must be support, neutral, or reject and should primarily reflect the
  substantive position.
- agreement: 1 = strongly disagree, 5 = strongly agree.
- legitimacy: 1 = completely illegitimate, 5 = completely legitimate from this
  persona's perspective.
- offensiveness: 1 = not offensive, 5 = extremely offensive.
- endorsement: 1 = would definitely not share, 5 = would definitely share the message
  as written.
- Keep stance and agreement coherent: support normally maps to 3..5, reject to 1..2,
  neutral to 2..3.
- emotional_reaction must contain at most 20 words.
- brief_reason must contain at most 60 words.
- Evaluate every message independently and return exactly one response per tweet_id.

[MESSAGES]
1. [tweet_id=<id>]
Message: <anonymised post>
  ... ten messages ...

[OUTPUT JSON SCHEMA]
{"responses": [{"tweet_id": "<id>", "stance": "support|neutral|reject", "agreement": 1,
  "legitimacy": 1, "offensiveness": 1, "endorsement": 1, "emotional_reaction": "short text",
  "brief_reason": "short text"}, ... ]}
\end{verbatim}
\caption{Prompt template. Angle brackets mark the parts filled in for each call; the indented alternatives show the two persona formats and the two kinds of condition line. Decoding: temperature 0, seed 42, 4,096 output tokens, JSON mode.}
\label{fig:appendix-prompt}
\end{figure*}

\subsection{Knowledge audit}
\label{app:audit}

Before the experiment, the model was queried 20 times about each of the three leaders and each of three control names: two invented but plausible Spanish names and one opaque code (Table~\ref{tab:appendix-audit}; 120 requests, using the same decoding settings as the experiment). For each name, the 20 requests comprised five semantically equivalent prompt variants in each of four independent probe families: recognition, party and political role, left--right ideology, and issue positions.  The ideology probes used a 1--10 scale, where 1 denotes the political left and 10 the political right. The issue-position probes used four further 1--10 scales: state--market orientation (1 = greater government intervention; 10 = greater market orientation and tax reduction), housing policy (1 = greater regulation and tenant protection; 10 = greater emphasis on housing supply, market mechanisms and home ownership), immigration restrictiveness (1 = a more open or humanitarian position; 10 = a more restrictive position emphasising border control), and institutional trust (1 = anti-establishment discourse or low institutional trust; 10 = institutional continuity or high institutional trust).  The model also differentiated the three leaders on several of these dimensions; for instance, their median immigration-restrictiveness scores were 3, 6 and 10. The ideological and policy-position scores describe the model's internal representations and are not treated as ground truth about the politicians. The audit checks whether the names activate stable and distinct political identities; it does not estimate the effect of speaker attribution.

\begin{table}[htbp]
\centering\footnotesize
\setlength{\tabcolsep}{3pt}
\begin{tabular}{lccc}
\toprule
\addlinespace[2pt]
Name & Recog. & Party & Left--right \\
\midrule
Pedro Sánchez & 5/5 & PSOE 5/5 & 3 (3--3) \\
Alberto Núñez Feijóo & 5/5 & PP 5/5 & 8 (7--8) \\
Santiago Abascal & 5/5 & VOX 5/5 & 9 (9--10) \\
Darío Ventosal Brumet & 0/5 & none & -- \\
Lucía Montalvo Cendral & 0/5 & none & -- \\
Opaque code & 0/5 & none & -- \\
\bottomrule
\end{tabular}
\caption{Knowledge audit based on 20 requests per name: five prompt variants for each of four probe families. Recognition and party identification are reported over the five relevant probes, while left--right placement is reported as the median (range) across the five ideology probes. The last three names are controls: two invented names and the code \texttt{ES\_POLITICIAN\_CONTROL\_QZ17}.}
\label{tab:appendix-audit}
\end{table}

\subsection{Role markers and implausible attributions}
\label{app:markers}

Anonymisation removes names, not positions. Table~\ref{tab:appendix-markers} lists the five groups of role markers that flag a post as implausible under another name, matched like the topic keywords (URLs and accents removed, lower case) and only in the posts of the authors each group applies to; regional and foreign governments and foreign prime ministers are removed before matching. Table~\ref{tab:appendix-flagged} gives the flagged stimuli by topic and real author. In an informal reading of random samples (not a formal annotation), precision was high: 54 of 60 marked posts read before the last fixes showed the author's position, the six errors were fixed, and 16 of 16 marked topic posts read afterwards did. Recall was low: most unmarked topic posts by Sánchez still spoke as the Government without a marker word (``vamos a destinar 2.275 M€''), and about half of those by Feijóo and Abascal still opposed it. The flags are therefore a lower bound. Filtering on them would leave 7 Abascal posts on household costs and 26 and 23 Feijóo and Abascal posts on corruption, and would still keep many implausible posts, so the design stays fully crossed. Table~\ref{tab:appendix-plausibility} reports the attribution effect by the correspondence between the displayed name and the real author; the comparison is descriptive, because the three categories hold different posts.

\begin{table*}[tbp]
\centering\small
\begin{tabular}{lll>{\raggedright\arraybackslash}p{6.8cm}}
\toprule
Marker group & Posts by & Implausible under & Example markers \\
\midrule
Government voice & Sánchez & Feijóo, Abascal & ``Gobierno'', ministers, Consejo de Ministros, EU presidency, summits, ``homólogo'' \\
Against the Government & Feijóo, Abascal & Sánchez & ``Gobierno'', ministers, ``Sánchez'', ``sanchismo'', resignation, motion of censure \\
Against the right & Sánchez & Feijóo, Abascal & ``ultraderecha'', ``la derecha'', ``reaccionarios'' \\
Against the left & Feijóo, Abascal & Sánchez & ``izquierda'', ``comunistas'', ``progres'' \\
Against the PP & Abascal & Feijóo & ``PPSOE'', ``bipartidismo'', ``derechita'', ``populares y socialistas'' \\
\bottomrule
\end{tabular}
\caption{Role-marker groups. A marker is matched only in the posts of the authors it applies to and flags the post as implausible under the names in the third column.}
\label{tab:appendix-markers}
\end{table*}

\begin{table*}[tbp]
\centering
\begin{minipage}[t]{0.44\textwidth}\vspace{0pt}
\centering\footnotesize
\setlength{\tabcolsep}{3pt}
\begin{tabular}{lrrrr}
\toprule
Topic & Sánchez & Feijóo & Abascal & All \\
\midrule
Economy & 83/304 & 33/99 & 10/30 & 126/433 \\
Household & 41/123 & 60/94 & 4/11 & 105/228 \\
Immigration & 17/50 & 22/40 & 19/56 & 58/146 \\
Corruption & 15/27 & 101/127 & 91/114 & 207/268 \\
\midrule
All & 156/504 & 216/360 & 124/211 & 496/1,075 \\
\bottomrule
\end{tabular}
\caption{Stimuli flagged by at least one role marker (flagged/all), by topic and real author: 31\% of Sánchez's, 60\% of Feijóo's and 59\% of Abascal's posts, 46\% in all.}
\label{tab:appendix-flagged}
\end{minipage}\hfill
\begin{minipage}[t]{0.54\textwidth}\vspace{0pt}
\centering\footnotesize
\setlength{\tabcolsep}{4pt}
\begin{tabular}{llrrr}
\toprule
Framework & Attribution & Pairs & $\Delta$ stance & 95\% CI \\
\midrule
IW & True author & 161,230 & -0.072 & [-0.084, -0.061] \\
IW & Plausible & 224,683 & -0.131 & [-0.144, -0.119] \\
IW & Implausible & 97,777 & -0.145 & [-0.161, -0.129] \\
\midrule
MFT & True author & 161,240 & -0.089 & [-0.106, -0.072] \\
MFT & Plausible & 224,683 & -0.149 & [-0.167, -0.132] \\
MFT & Implausible & 97,797 & -0.233 & [-0.260, -0.205] \\
\bottomrule
\end{tabular}
\caption{Speaker-attribution effects on stance by the correspondence between the displayed name and the real author. A counterfactual attribution is implausible when a role marker flags the post under that name.}
\label{tab:appendix-plausibility}
\end{minipage}
\end{table*}

\subsection{Completeness}
\label{app:completeness}

Table~\ref{tab:appendix-completeness} gives the persona--post pairs observed in all four conditions by framework and topic. Completion exceeded 99.95\% in every cell; the missing quartets come from calls whose JSON failed validation after the retries.

\begin{table*}[tbp]
\centering\footnotesize
\setlength{\tabcolsep}{4pt}
\begin{tabular}{lrrrrrrrr}
\toprule
 & \multicolumn{4}{c}{IW} & \multicolumn{4}{c}{MFT} \\
\cmidrule(lr){2-5}\cmidrule(lr){6-9}
 & Economy & Household & Immigration & Corruption & Economy & Household & Immigration & Corruption \\
\midrule
Pairs & 64,950 & 34,200 & 21,900 & 40,200 & 64,950 & 34,200 & 21,900 & 40,200 \\
Complete & 64,950 & 34,200 & 21,890 & 40,190 & 64,940 & 34,200 & 21,900 & 40,200 \\
\% & 100.00 & 100.00 & 99.95 & 99.95 & 99.99 & 100.00 & 100.00 & 100.00 \\
\bottomrule
\end{tabular}
\caption{Complete four-condition quartets by framework and topic.}
\label{tab:appendix-completeness}
\end{table*}

\section{Personas and Ideology}
\label{app:personas}

\subsection{Persona selection}
\label{app:selection}

The persona pool is drawn from the resource of \citet{greco-lacava-tagarelli-2026-culturally}. It contains profiles generated from ten World Values Survey variables---religiosity, child-rearing values, moral acceptability, trust and social capital, political participation, national pride, happiness, gender equality, materialism orientation and attitudes towards outgroups---together with a biography, Inglehart--Welzel coordinates and MFQ-2 scores. 
For each topic and theoretical framework, we defined high, mid and low target profiles in the corresponding value space (Table~\ref{tab:appendix-profiles}). 
We followed the same persona-ranking and selection procedure used by \citet{greco-etal-2026-morality} in their study of the reception of Trump's political messages, adapting only the mapping between the political topics and the relevant IW dimensions or moral foundations. We retained the 50 highest-ranked personas for each topic--profile combination, without reusing a persona across levels of the same topic. High, mid and low describe positions in the corresponding theoretical value space and are never shown to the model in the prompt.  

\begin{table*}[tbp]
\centering\small
\begin{tabular}{ll>{\raggedright\arraybackslash}p{5.3cm}>{\raggedright\arraybackslash}p{5.3cm}}
\toprule
Topic & Level & IW profile & MFT profile \\
\midrule
Economy, employment & high & material security with a weak rational-policy component & merit and economic contribution \\
 & mid & central profile without a strong topic direction & no strongly polarised economic orientation \\
 & low & self-expression and quality-of-work orientation & redistribution and social protection \\
\addlinespace
Household costs & high & affordability and basic material security & affordability and protection of basic needs \\
 & mid & central profile without a strong topic direction & no strongly polarised household-cost orientation \\
 & low & residential choice, autonomy and self-expression & market allocation and individual responsibility \\
\addlinespace
Immigration, border & high & inclusive, open and pluralist orientation & humanitarian inclusion \\
 & mid & central profile without a strong topic direction & no strongly polarised immigration orientation \\
 & low & restriction, security and traditional order & restriction and border control \\
\addlinespace
Corruption & high & procedural and institutional accountability & strong institutional-accountability sensitivity \\
 & mid & central profile without a strong topic direction & moderate institutional-accountability sensitivity \\
 & low & traditional or personalised authority orientation & weak institutional-accountability sensitivity (not support for corruption) \\
\bottomrule
\end{tabular}
\caption{Topic-conditioned profile levels used to select 50 personas per cell in each framework. The levels order personas within the framework's value space; they are not ideological labels.}
\label{tab:appendix-profiles}
\end{table*}

\begin{figure*}[htbp]
    \centering
    \includegraphics[width=0.7\textwidth]
    {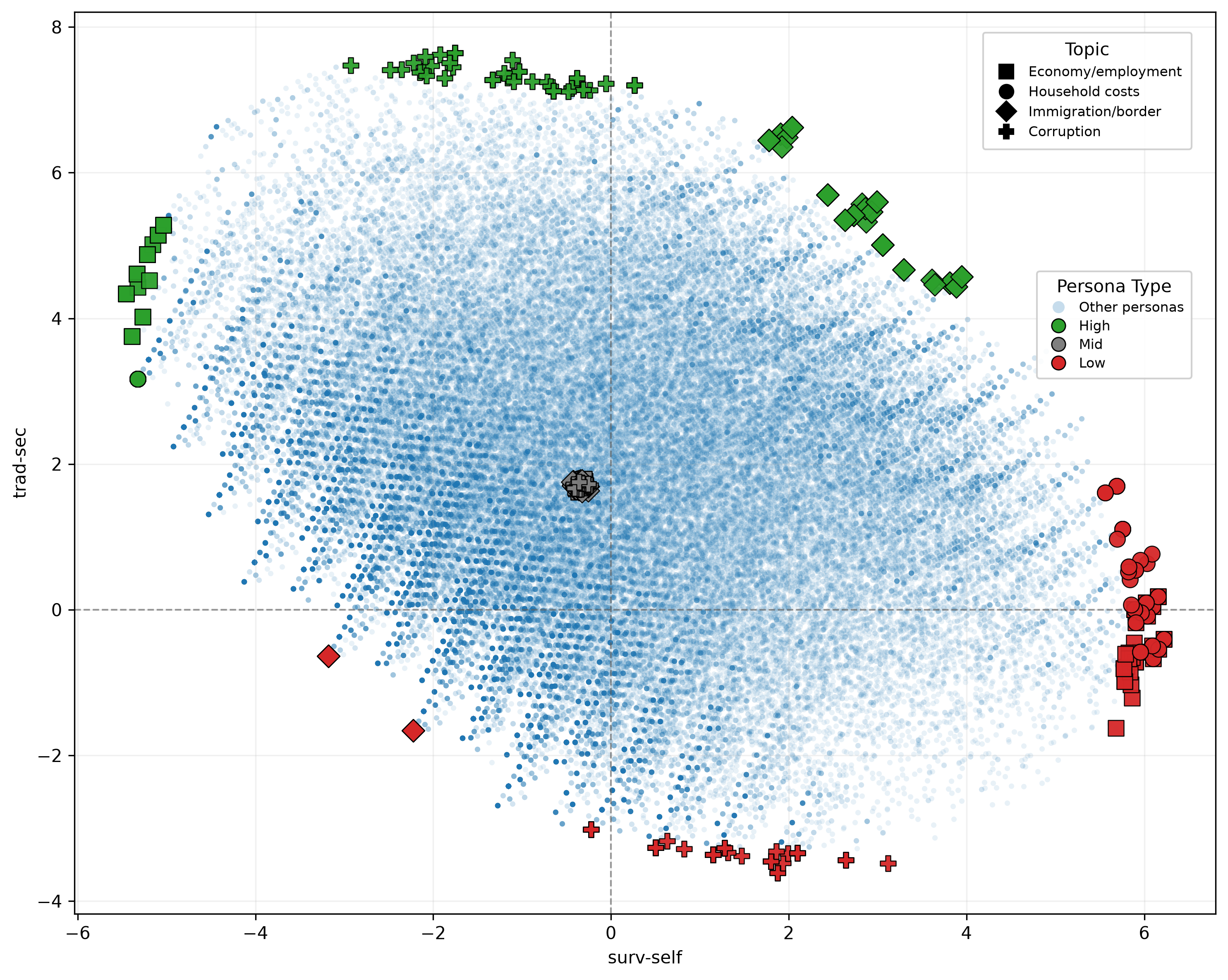}
    \caption{Inglehart--Welzel personas selected for each
    topic--profile pair. Background points show the complete persona
    pool in the IW space, while highlighted markers show the 50 selected
    personas for each topic and profile. Colours denote the
    topic-conditioned profiles (\textit{high}, \textit{mid} and
    \textit{low}), and marker shapes denote topics.}
    \label{fig:iw-selected-personas}
\end{figure*}

Figure~\ref{fig:iw-selected-personas} shows the resulting IW selections relative to the complete persona pool. Table~\ref{tab:mft-selected-profiles-k50} summarises the corresponding MFT selections through the mean and standard deviation of each moral-foundation score across the 50 selected personas.

\begin{table*}[htbp]
\centering
\small
\setlength{\tabcolsep}{4pt}
\begin{tabular}{lcccccc}
\toprule
Profile & Care & Equality & Proportionality & Loyalty & Authority & Purity \\
\midrule
\multicolumn{7}{l}{\textbf{Economy/employment}} \\
High & \cellcolor{gray!3} 1.17 $\pm$ 0.21 & \cellcolor{gray!3} 1.20 $\pm$ 0.13 & \cellcolor{gray!45} 4.79 $\pm$ 0.13 & \cellcolor{gray!19} 2.71 $\pm$ 0.61 & \cellcolor{gray!33} 3.47 $\pm$ 0.47 & \cellcolor{gray!17} 2.43 $\pm$ 0.50 \\
Mid & \cellcolor{gray!27} 3.26 $\pm$ 0.43 & \cellcolor{gray!19} 2.73 $\pm$ 0.35 & \cellcolor{gray!33} 3.45 $\pm$ 0.35 & \cellcolor{gray!22} 2.87 $\pm$ 0.39 & \cellcolor{gray!25} 3.11 $\pm$ 0.37 & \cellcolor{gray!22} 2.80 $\pm$ 0.26 \\
Low & \cellcolor{gray!60} 5.00 $\pm$ 0.02 & \cellcolor{gray!50} 4.70 $\pm$ 0.32 & \cellcolor{gray!16} 2.32 $\pm$ 0.38 & \cellcolor{gray!23} 2.93 $\pm$ 0.43 & \cellcolor{gray!24} 2.99 $\pm$ 0.55 & \cellcolor{gray!23} 2.92 $\pm$ 0.64 \\
\midrule
\multicolumn{7}{l}{\textbf{Household costs}} \\
High & \cellcolor{gray!60} 5.00 $\pm$ 0.02 & \cellcolor{gray!50} 4.70 $\pm$ 0.32 & \cellcolor{gray!16} 2.32 $\pm$ 0.38 & \cellcolor{gray!23} 2.93 $\pm$ 0.43 & \cellcolor{gray!24} 2.99 $\pm$ 0.55 & \cellcolor{gray!23} 2.92 $\pm$ 0.64 \\
Mid & \cellcolor{gray!27} 3.26 $\pm$ 0.43 & \cellcolor{gray!19} 2.73 $\pm$ 0.35 & \cellcolor{gray!33} 3.45 $\pm$ 0.35 & \cellcolor{gray!22} 2.87 $\pm$ 0.39 & \cellcolor{gray!25} 3.11 $\pm$ 0.37 & \cellcolor{gray!22} 2.80 $\pm$ 0.26 \\
Low & \cellcolor{gray!3} 1.17 $\pm$ 0.21 & \cellcolor{gray!3} 1.20 $\pm$ 0.13 & \cellcolor{gray!45} 4.79 $\pm$ 0.13 & \cellcolor{gray!19} 2.71 $\pm$ 0.61 & \cellcolor{gray!33} 3.47 $\pm$ 0.47 & \cellcolor{gray!17} 2.43 $\pm$ 0.50 \\
\midrule
\multicolumn{7}{l}{\textbf{Immigration/border}} \\
High & \cellcolor{gray!45} 4.86 $\pm$ 0.29 & \cellcolor{gray!24} 3.01 $\pm$ 0.48 & \cellcolor{gray!30} 3.49 $\pm$ 0.45 & \cellcolor{gray!7} 1.93 $\pm$ 0.21 & \cellcolor{gray!5} 1.80 $\pm$ 0.27 & \cellcolor{gray!16} 2.27 $\pm$ 0.42 \\
Mid & \cellcolor{gray!27} 3.26 $\pm$ 0.43 & \cellcolor{gray!19} 2.73 $\pm$ 0.35 & \cellcolor{gray!33} 3.45 $\pm$ 0.35 & \cellcolor{gray!22} 2.87 $\pm$ 0.39 & \cellcolor{gray!25} 3.11 $\pm$ 0.37 & \cellcolor{gray!22} 2.80 $\pm$ 0.26 \\
Low & \cellcolor{gray!5} 1.77 $\pm$ 0.47 & \cellcolor{gray!7} 1.91 $\pm$ 0.56 & \cellcolor{gray!38} 4.25 $\pm$ 0.41 & \cellcolor{gray!43} 4.67 $\pm$ 0.24 & \cellcolor{gray!45} 4.79 $\pm$ 0.28 & \cellcolor{gray!26} 3.17 $\pm$ 0.51 \\
\midrule
\multicolumn{7}{l}{\textbf{Corruption}} \\
High & \cellcolor{gray!37} 4.14 $\pm$ 0.98 & \cellcolor{gray!37} 4.09 $\pm$ 0.45 & \cellcolor{gray!42} 4.57 $\pm$ 0.39 & \cellcolor{gray!33} 3.42 $\pm$ 0.64 & \cellcolor{gray!42} 4.53 $\pm$ 0.38 & \cellcolor{gray!31} 3.57 $\pm$ 0.64 \\
Mid & \cellcolor{gray!27} 3.26 $\pm$ 0.43 & \cellcolor{gray!19} 2.73 $\pm$ 0.35 & \cellcolor{gray!33} 3.45 $\pm$ 0.35 & \cellcolor{gray!22} 2.87 $\pm$ 0.39 & \cellcolor{gray!25} 3.11 $\pm$ 0.37 & \cellcolor{gray!22} 2.80 $\pm$ 0.26 \\
Low & \cellcolor{gray!33} 3.40 $\pm$ 0.80 & \cellcolor{gray!7} 1.92 $\pm$ 0.44 & \cellcolor{gray!26} 3.16 $\pm$ 0.32 & \cellcolor{gray!14} 2.14 $\pm$ 0.49 & \cellcolor{gray!13} 2.06 $\pm$ 0.36 & \cellcolor{gray!13} 2.08 $\pm$ 0.40 \\
\bottomrule
\end{tabular}
\caption{ MFQ-2 foundation scores (1--5) among the 50 personas selected for each topic-conditioned profile. Values are reported as mean $\pm$ standard deviation. Cell shading is proportional to the mean score: lighter gray indicates lower values and darker gray indicates higher values.}
\label{tab:mft-selected-profiles-k50}
\end{table*}

\subsection{Elicited self-placement}
\label{app:ideology}

Each of the 791 distinct personas (441 IW, 350 MFT) answered once, with the system prompt and persona profile of the reaction prompts followed by the question: \textit{In general, when talking about politics in Spain, the expressions ``left'' and ``right'' are normally used. On a scale from 1 to 10, where 1 means ``the furthest to the left'' and 10 means ``the furthest to the right'' of Spanish politics, where would you place yourself?} This is an English adaptation of the CIS barometer item. The scale is anchored to Spanish politics because most personas are not Spanish, and the personas are not presented as residents of Spain, as in the reaction prompts. Parties and leaders are not mentioned, because leader preferences are the outcome compared in RQ4. The settings were those of the reactions, and all 791 answers were valid integers from 1 to 10 (Table~\ref{tab:appendix-ideology-distribution}). Many IW personas placed themselves at 5, the lower of the two central categories on the 1--10 scale: 180 of 441 (41\%) answered exactly 5, compared with 59 of 350 MFT personas (17\%) and 22\% of respondents in the October 2026 barometer.

\begin{table*}[tbp]
\centering
\begin{minipage}[t]{0.48\textwidth}\vspace{0pt}
\centering\footnotesize
\setlength{\tabcolsep}{3pt}
\begin{tabular}{lrrrr}
\toprule
& Left & Centre & Right & Mean \\
\midrule
IW personas (441) & 27.2\% & 44.4\% & 28.3\% & 5.40 \\
MFT personas (350) & 39.4\% & 25.7\% & 34.9\% & 5.36 \\
CIS, Oct.\ 2026 (5,024) & 45.4\% & 29.2\% & 22.0\% & 4.51 \\
\bottomrule\end{tabular}
\caption{Elicited left--right self-placement of the personas and of CIS respondents \citep{cis-2026-3582}. CIS shares are of all respondents; 3.4\% answered ``don't know'' or did not answer, an option the personas did not have.}
\label{tab:appendix-ideology-distribution}
\end{minipage}\hfill
\begin{minipage}[t]{0.48\textwidth}\vspace{0pt}
\centering\footnotesize
\setlength{\tabcolsep}{4pt}
\begin{tabular}{lrc}
\toprule
Value dimension & $\rho$ & 95\% CI \\
\midrule
\multicolumn{3}{l}{\textit{IW personas}} \\
Survival/self-expression & $-0.60$ & $[-0.66, -0.54]$ \\
Traditional/secular-rational & $-0.06$ & $[-0.16, 0.03]$ \\
\addlinespace
\multicolumn{3}{l}{\textit{MFT personas}} \\
Care & $-0.87$ & $[-0.89, -0.84]$ \\
Equality & $-0.67$ & $[-0.73, -0.61]$ \\
Proportionality & $0.67$ & $[0.61, 0.73]$ \\
Loyalty & $0.43$ & $[0.34, 0.51]$ \\
Authority & $0.59$ & $[0.52, 0.66]$ \\
Purity & $0.21$ & $[0.10, 0.31]$ \\
\bottomrule
\end{tabular}
\caption{Spearman correlation between the elicited self-placement (1--10) and each value dimension shown in the persona prompts, one observation per persona (441 IW, 350 MFT). Negative values mean that higher scores go with placements further to the left. Confidence intervals from Fisher's $z$ with variance $1.06/(n-3)$ \citep{fieller-etal-1957-tests}.}
\label{tab:appendix-values-correlations}
\end{minipage}
\end{table*}

\subsection{Values and self-placement}
\label{app:values}

Table~\ref{tab:appendix-values-groups} gives the ideological groups at each profile level, and Table~\ref{tab:appendix-values-correlations} the correlation between the self-placement and each value dimension shown in the prompts. Where a profile has a clear political direction, it was strongly associated with the elicited ideological group: the inclusive and restrictive immigration profiles were mostly left- and right-leaning in both frameworks, and the MFT economy and household-cost profiles, which contrast care and equality with proportionality, fell entirely on one side. Corruption profiles did not produce a clear monotonic separation by ideology in either framework, and mid levels were mostly centre. Among MFT personas, the placement followed the foundation scores: further left with care and equality, further right with proportionality, authority and loyalty, and only weakly with purity. These are the directions found by \citet{atari-etal-2023-morality} in samples from 19 countries that answered a similar 10-point left--right item: right-wing placement was associated negatively with care and equality and positively with proportionality, loyalty, authority and purity, with purity the weakest association. Those estimates come from a model with all six foundations together, so only the directions are comparable. Among IW personas, the placement followed the survival--self-expression coordinate but hardly the traditional--secular-rational one. Because some MFT cells hold the same personas (Appendix~\ref{app:selection}), their fully one-sided splits mainly show that the model reads high care and equality as left and high proportionality as right, and the profile-level heterogeneity of Figure~\ref{fig:appendix-profile-heterogeneity} and the ideological groups of RQ2 are not independent.

\begin{table*}[tbp]
\centering\footnotesize
\setlength{\tabcolsep}{3pt}
\begin{tabular}{ll>{\raggedright\arraybackslash}p{3.3cm}rrr>{\raggedright\arraybackslash}p{3.3cm}rrr}
\toprule
& & \multicolumn{4}{c}{IW personas} & \multicolumn{4}{c}{MFT personas} \\
\cmidrule(lr){3-6}\cmidrule(lr){7-10}
Topic & Level & Profile & L & C & R & Profile & L & C & R \\
\midrule
Immigration & high & Inclusive, pluralist & 82 & 16 & 2 & Humanitarian inclusion & 88 & 12 & 0 \\
 & mid & Central profile & 16 & 54 & 30 & Not polarised & 22 & 72 & 6 \\
 & low & Restriction, traditional order & 0 & 36 & 64 & Restriction, border control & 0 & 0 & 100 \\
\addlinespace
Economy & high & Material security & 0 & 70 & 30 & Merit, contribution & 0 & 0 & 100 \\
 & mid & Central profile & 14 & 62 & 24 & Not polarised & 22 & 72 & 6 \\
 & low & Self-expression, quality of work & 74 & 26 & 0 & Redistribution, social protection & 100 & 0 & 0 \\
\addlinespace
Household costs & high & Affordability, material security & 0 & 44 & 56 & Affordability, basic needs & 100 & 0 & 0 \\
 & mid & Central profile & 12 & 58 & 30 & Not polarised & 22 & 72 & 6 \\
 & low & Autonomy, self-expression & 74 & 24 & 2 & Market, individual responsibility & 0 & 0 & 100 \\
\addlinespace
Corruption & high & Institutional accountability & 10 & 72 & 18 & Strong accountability sensitivity & 36 & 30 & 34 \\
 & mid & Central profile & 16 & 54 & 30 & Moderate accountability sensitivity & 22 & 72 & 6 \\
 & low & Personalised authority & 20 & 48 & 32 & Weak accountability sensitivity & 30 & 66 & 4 \\
\bottomrule
\end{tabular}
\caption{Ideological groups of the personas selected at each topic-conditioned profile level: share (\%) of the 50 personas who placed themselves on the left (L, 1--4), centre (C, 5--6) or right (R, 7--10). A persona selected for several topics counts in each. Profile descriptions are abbreviated from Table~\ref{tab:appendix-profiles}.}
\label{tab:appendix-values-groups}
\end{table*}

\subsection{Framework, persona rank and profile level}
\label{app:sensitivity}

\begin{figure*}[tbp]
\centering
\includegraphics[width=\textwidth]{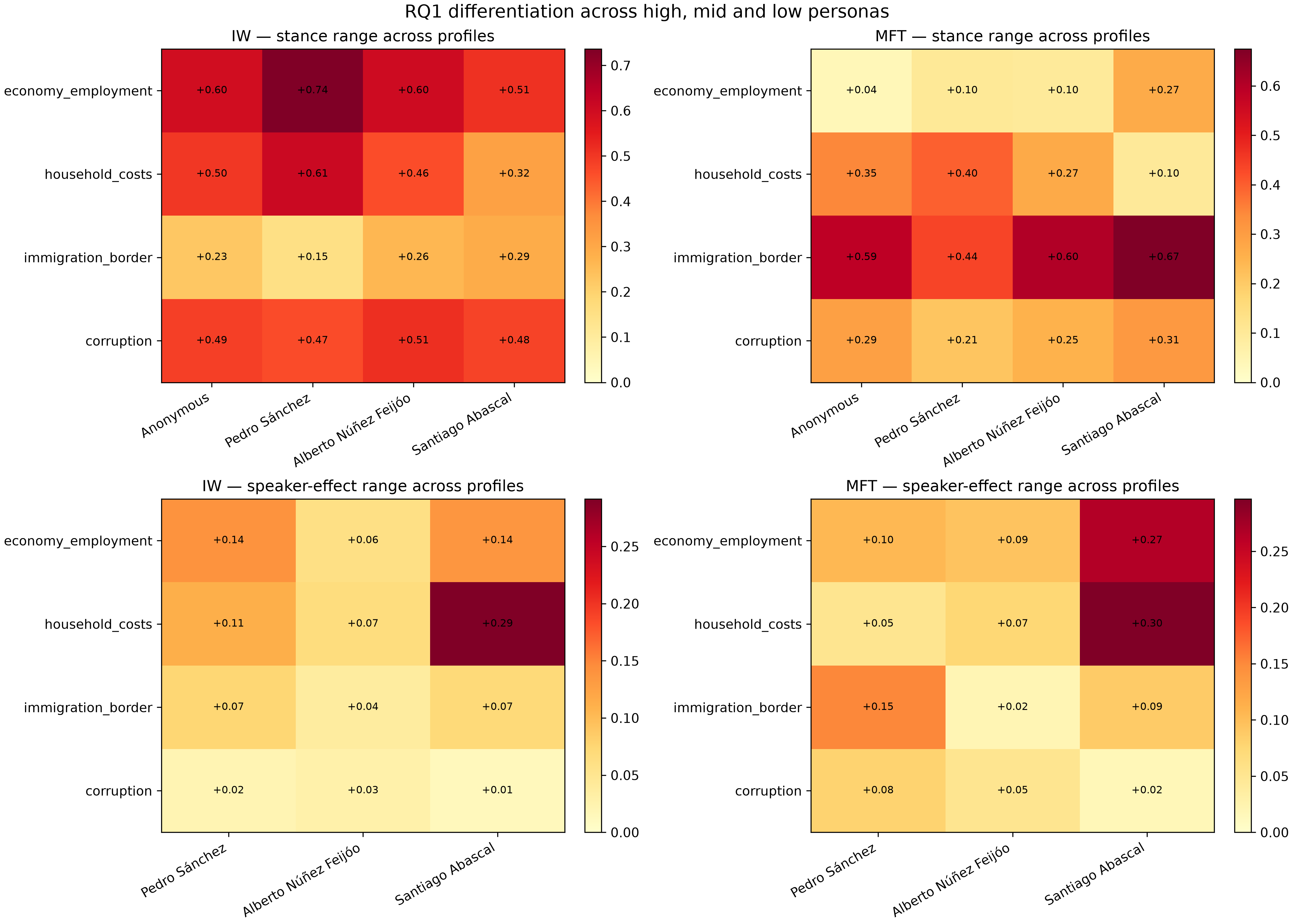}
\caption{Variation across the high, mid and low profile levels. Upper panels: for each topic and condition, the range (maximum minus minimum) of the mean stance across the three levels. Lower panels: the range of the speaker-attribution effect. Left panels IW, right panels MFT. The levels are value profiles, not ideology.}
\label{fig:appendix-profile-heterogeneity}
\end{figure*}

The attribution penalty appears under both selection frameworks, with somewhat larger Sánchez and Feijóo effects among MFT personas (Table~\ref{tab:appendix-framework}). The pattern is robust to the number of personas retained per cell: all nested cohorts, from the best-ranked persona to the full top-50, produce negative attribution effects, and Abascal consistently receives the largest penalty. The relative ordering of Sánchez and Feijóo varies slightly across the IW cohorts, but their effects remain close (Table~\ref{tab:appendix-rank-sensitivity}). Figure~\ref{fig:appendix-profile-heterogeneity} shows how much the mean stance and the attribution effect vary across the high, mid and low profile levels. The levels are topic-conditioned value profiles, not left--right ideology, so this heterogeneity must not be read as evidence on RQ2.

\begin{table*}[tbp]
\centering
\begin{minipage}[t]{0.48\textwidth}\vspace{0pt}
\centering\footnotesize
\setlength{\tabcolsep}{4pt}
\begin{tabular}{llrr}
\toprule
Framework & Speaker & $\Delta$ stance & 95\% CI \\
\midrule
IW & Sánchez & -0.079 & [-0.092, -0.066] \\
IW & Feijóo & -0.081 & [-0.093, -0.070] \\
IW & Abascal & -0.183 & [-0.200, -0.166] \\
\midrule
MFT & Sánchez & -0.122 & [-0.143, -0.101] \\
MFT & Feijóo & -0.118 & [-0.136, -0.100] \\
MFT & Abascal & -0.198 & [-0.220, -0.177] \\
\bottomrule\end{tabular}
\caption{Speaker-attribution effects on stance by persona-selection framework, all topics.}
\label{tab:appendix-framework}
\end{minipage}\hfill
\begin{minipage}[t]{0.48\textwidth}\vspace{0pt}
\centering\footnotesize
\setlength{\tabcolsep}{2.5pt}
\begin{tabular}{llrrr}
\toprule
Outcome & Framework & Effect range & Neg. & Pos. \\
\midrule
Agreement & IW & [-0.365, -0.076] & 12/12 & 0/12 \\
Agreement & MFT & [-0.676, -0.031] & 12/12 & 0/12 \\
Endorsement & IW & [-0.495, -0.087] & 12/12 & 0/12 \\
Endorsement & MFT & [-0.585, -0.041] & 12/12 & 0/12 \\
Legitimacy & IW & [-0.334, -0.048] & 12/12 & 0/12 \\
Legitimacy & MFT & [-0.524, -0.024] & 12/12 & 0/12 \\
Offensiveness & IW & [0.003, 0.316] & 0/12 & 12/12 \\
Offensiveness & MFT & [0.014, 0.433] & 0/12 & 12/12 \\
\bottomrule
\end{tabular}
\caption{Speaker-attribution effects on the 1--5 ratings: range of the mean effect over the twelve topic--speaker cells of each framework, and the number of cells with a negative or positive effect.}
\label{tab:appendix-secondary}
\end{minipage}
\end{table*}

\begin{table*}[tbp]
\centering\small
\begin{tabular}{llccc}
\toprule
Framework & Persona cohort & Sánchez & Feijóo & Abascal \\
\midrule
IW & Best & -0.115 [-0.204, -0.026] & -0.067 [-0.122, -0.012] & -0.178 [-0.263, -0.094] \\
IW & Top-5 & -0.094 [-0.132, -0.056] & -0.078 [-0.106, -0.049] & -0.173 [-0.214, -0.132] \\
IW & Top-20 & -0.075 [-0.094, -0.057] & -0.084 [-0.101, -0.067] & -0.181 [-0.207, -0.155] \\
IW & Top-50 & -0.079 [-0.092, -0.066] & -0.081 [-0.093, -0.070] & -0.183 [-0.200, -0.166] \\
\midrule
MFT & Best & -0.124 [-0.234, -0.014] & -0.123 [-0.229, -0.016] & -0.204 [-0.324, -0.084] \\
MFT & Top-5 & -0.124 [-0.173, -0.074] & -0.114 [-0.152, -0.076] & -0.173 [-0.214, -0.132] \\
MFT & Top-20 & -0.119 [-0.147, -0.092] & -0.116 [-0.140, -0.092] & -0.199 [-0.232, -0.166] \\
MFT & Top-50 & -0.122 [-0.143, -0.101] & -0.118 [-0.136, -0.100] & -0.198 [-0.220, -0.177] \\
\bottomrule\end{tabular}
\caption{Speaker-attribution effects on stance with nested persona cohorts (the best-ranked persona per cell, the top 5, the top 20 and all 50). Cells report the mean effect [95\% CI].}
\label{tab:appendix-rank-sensitivity}
\end{table*}

\section{Additional Results}
\label{app:results}

\subsection{Speaker attribution}
\label{app:rq1}

\begin{figure*}[tbp]
\centering
\includegraphics[width=\textwidth]{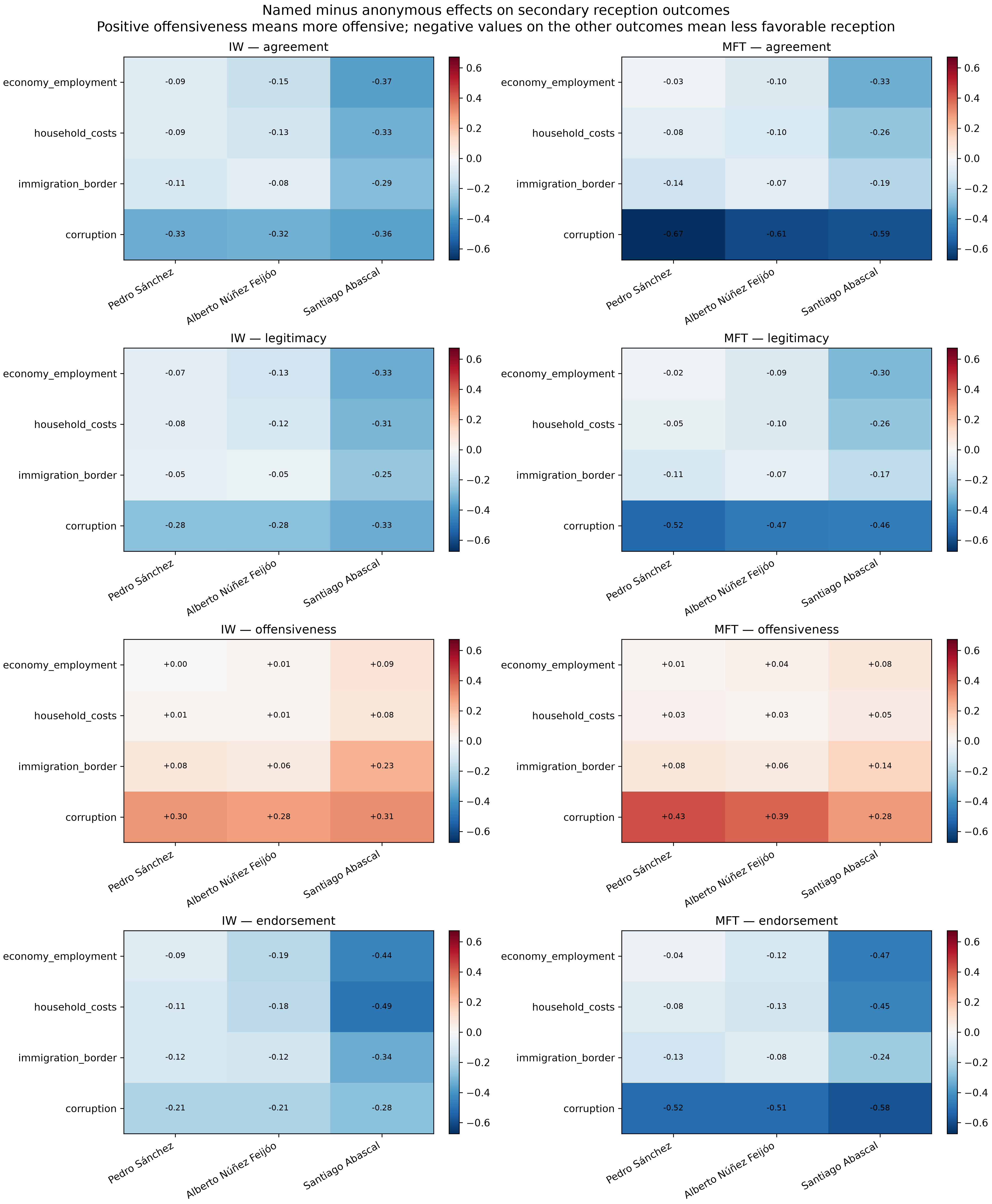}
\caption{Speaker-attribution effects on agreement, legitimacy, offensiveness and endorsement by topic, speaker and framework.}
\label{fig:appendix-secondary-outcomes}
\end{figure*}

Table~\ref{tab:appendix-secondary} and Figure~\ref{fig:appendix-secondary-outcomes} give the speaker-attribution effects on the four ratings: in every framework--topic--speaker cell, naming the leader lowered agreement, legitimacy and endorsement and raised perceived offensiveness. Table~\ref{tab:appendix-topic-ci} gives the confidence intervals of the topic estimates of Table~\ref{tab:rq3-topics}.

\begin{table*}[tbp]
\centering\small
\begin{tabular}{llccc}
\toprule
Framework & Topic & Sánchez & Feijóo & Abascal \\
\midrule
IW & Economy and employment & -0.030 [-0.046, -0.014] & -0.055 [-0.070, -0.040] & -0.186 [-0.210, -0.161] \\
IW & Household costs & -0.051 [-0.072, -0.031] & -0.057 [-0.078, -0.036] & -0.168 [-0.207, -0.130] \\
IW & Immigration and border & -0.056 [-0.077, -0.034] & -0.030 [-0.052, -0.007] & -0.164 [-0.196, -0.131] \\
IW & Corruption & -0.194 [-0.219, -0.168] & -0.173 [-0.197, -0.149] & -0.202 [-0.233, -0.170] \\
\midrule
MFT & Economy and employment & -0.008 [-0.024, 0.008] & -0.042 [-0.060, -0.025] & -0.182 [-0.212, -0.152] \\
MFT & Household costs & -0.053 [-0.075, -0.032] & -0.047 [-0.064, -0.029] & -0.132 [-0.165, -0.098] \\
MFT & Immigration and border & -0.065 [-0.090, -0.040] & -0.031 [-0.049, -0.012] & -0.101 [-0.130, -0.072] \\
MFT & Corruption & -0.395 [-0.436, -0.355] & -0.348 [-0.383, -0.312] & -0.335 [-0.376, -0.295] \\
\bottomrule\end{tabular}
\caption{Speaker-attribution effects on stance by topic and framework. Cells report the mean effect [95\% CI].}
\label{tab:appendix-topic-ci}
\end{table*}

\subsection{Ideological congruence}
\label{app:rq2}

Tables~\ref{tab:appendix-rq2-groups} to~\ref{tab:appendix-rq2-outcomes} give the RQ2 estimates by group, by framework and group, by framework and topic, by topic and group, and by outcome; Table~\ref{tab:appendix-rq2-placement} gives the effects at each placement. Agreement, legitimacy and endorsement show the same congruence pattern as stance, with larger values because they use a 1--5 scale; offensiveness shows it with the opposite sign, since a negative contrast means that the left group found A's messages less offensive, relative to B's, than the right group did.

\begin{table*}[tbp]
\centering\footnotesize
\setlength{\tabcolsep}{5pt}
\begin{tabular}{lrrrrrrrrrr}
\toprule
Placement & 1 & 2 & 3 & 4 & 5 & 6 & 7 & 8 & 9 & 10 \\
\midrule
Personas & 1 & 63 & 129 & 65 & 239 & 47 & 40 & 152 & 41 & 14 \\
Sánchez & -0.002 & 0.001 & -0.046 & -0.097 & -0.114 & -0.154 & -0.124 & -0.126 & -0.151 & -0.140 \\
Feijóo & -0.145 & -0.110 & -0.112 & -0.129 & -0.120 & -0.125 & -0.080 & -0.048 & 0.008 & -0.046 \\
Abascal & -0.328 & -0.342 & -0.270 & -0.263 & -0.194 & -0.198 & -0.109 & -0.064 & -0.037 & -0.104 \\
\bottomrule
\end{tabular}
\caption{Speaker-attribution effects on stance at each elicited left--right placement, both frameworks and all topics (the values plotted in Figure~\ref{fig:placement}a; confidence intervals in the figure). Placement 1 holds one persona.}
\label{tab:appendix-rq2-placement}
\end{table*}

\begin{table*}[tbp]
\centering\small
\setlength{\tabcolsep}{4pt}
\begin{tabular}{lrccc}
\toprule
Persona group & Personas & Sánchez & Feijóo & Abascal \\
\midrule
Left (1--4) & 258 & -0.047 [-0.066, -0.027] & -0.116 [-0.133, -0.099] & -0.288 [-0.313, -0.263] \\
Centre (5--6) & 286 & -0.123 [-0.141, -0.105] & -0.121 [-0.137, -0.105] & -0.195 [-0.212, -0.178] \\
Right (7--10) & 247 & -0.129 [-0.149, -0.109] & -0.049 [-0.067, -0.030] & -0.072 [-0.090, -0.053] \\
\midrule
All & 791 & -0.100 [-0.114, -0.086] & -0.100 [-0.112, -0.087] & -0.191 [-0.206, -0.175] \\
\bottomrule
\end{tabular}
\caption{Speaker-attribution effects on stance by elicited ideological group, both frameworks and all topics. Cells report the mean effect [95\% CI, two-way cluster-robust by message and persona].}
\label{tab:appendix-rq2-groups}
\end{table*}

\begin{table*}[tbp]
\centering\small
\setlength{\tabcolsep}{3pt}
\begin{tabular}{llrccc}
\toprule
Framework & Group & Personas & Sánchez & Feijóo & Abascal \\
\midrule
IW & Left & 120 & -0.023 [-0.043, -0.003] & -0.091 [-0.108, -0.074] & -0.291 [-0.320, -0.261] \\
IW & Centre & 196 & -0.081 [-0.097, -0.066] & -0.089 [-0.104, -0.074] & -0.169 [-0.189, -0.150] \\
IW & Right & 125 & -0.134 [-0.158, -0.110] & -0.056 [-0.080, -0.031] & -0.096 [-0.122, -0.070] \\
\midrule
MFT & Left & 138 & -0.063 [-0.091, -0.036] & -0.133 [-0.157, -0.109] & -0.286 [-0.318, -0.254] \\
MFT & Centre & 90 & -0.187 [-0.218, -0.156] & -0.170 [-0.196, -0.143] & -0.234 [-0.259, -0.210] \\
MFT & Right & 122 & -0.125 [-0.152, -0.098] & -0.043 [-0.068, -0.018] & -0.052 [-0.075, -0.030] \\
\bottomrule\end{tabular}
\caption{Speaker-attribution effects on stance by framework and elicited ideological group, all topics. Cells report the mean effect [95\% CI].}
\label{tab:appendix-rq2-framework}
\end{table*}

\begin{table*}[tbp]
\centering\small
\setlength{\tabcolsep}{3pt}
\begin{tabular}{llccc}
\toprule
Framework & Topic & Sánchez -- Feijóo & Sánchez -- Abascal & Feijóo -- Abascal \\
\midrule
Both & All topics & 0.150 [0.128, 0.172] & 0.298 [0.264, 0.333] & 0.148 [0.124, 0.173] \\
Both & Economy & 0.175 [0.143, 0.208] & 0.321 [0.274, 0.368] & 0.146 [0.116, 0.176] \\
Both & Household costs & 0.143 [0.104, 0.182] & 0.366 [0.305, 0.427] & 0.223 [0.177, 0.268] \\
Both & Immigration & 0.150 [0.114, 0.186] & 0.196 [0.143, 0.249] & 0.045 [0.003, 0.088] \\
Both & Corruption & 0.081 [0.032, 0.130] & 0.210 [0.148, 0.272] & 0.128 [0.072, 0.184] \\
\midrule
IW & All topics & 0.147 [0.118, 0.176] & 0.306 [0.261, 0.351] & 0.159 [0.123, 0.195] \\
IW & Economy & 0.159 [0.118, 0.200] & 0.286 [0.229, 0.344] & 0.127 [0.080, 0.175] \\
IW & Household costs & 0.170 [0.111, 0.229] & 0.426 [0.339, 0.513] & 0.256 [0.187, 0.326] \\
IW & Immigration & 0.143 [0.096, 0.190] & 0.165 [0.099, 0.230] & 0.022 [-0.038, 0.081] \\
IW & Corruption & 0.056 [-0.008, 0.119] & 0.205 [0.122, 0.289] & 0.150 [0.077, 0.222] \\
\midrule
MFT & All topics & 0.152 [0.124, 0.180] & 0.295 [0.250, 0.339] & 0.143 [0.113, 0.173] \\
MFT & Economy & 0.187 [0.147, 0.226] & 0.342 [0.284, 0.400] & 0.155 [0.121, 0.189] \\
MFT & Household costs & 0.123 [0.081, 0.166] & 0.325 [0.258, 0.391] & 0.201 [0.149, 0.253] \\
MFT & Immigration & 0.157 [0.108, 0.206] & 0.224 [0.159, 0.289] & 0.067 [0.020, 0.115] \\
MFT & Corruption & 0.121 [0.044, 0.197] & 0.278 [0.200, 0.356] & 0.157 [0.071, 0.243] \\
\bottomrule\end{tabular}
\caption{Congruence contrasts on stance, $(\Delta_A-\Delta_B)_{\text{left}}-(\Delta_A-\Delta_B)_{\text{right}}$, by framework and topic. Positive values indicate congruence. Cells report the estimate [95\% CI].}
\label{tab:appendix-rq2-congruence-topic}
\end{table*}

\begin{table*}[tbp]
\centering\small
\setlength{\tabcolsep}{3pt}
\begin{tabular}{llrccc}
\toprule
Topic & Group & Personas & Sánchez & Feijóo & Abascal \\
\midrule
Economy & Left & 105 & 0.040 [0.026, 0.055] & -0.077 [-0.098, -0.057] & -0.287 [-0.322, -0.251] \\
Economy & Centre & 115 & -0.043 [-0.060, -0.026] & -0.053 [-0.069, -0.037] & -0.170 [-0.193, -0.147] \\
Economy & Right & 80 & -0.063 [-0.084, -0.042] & -0.005 [-0.025, 0.015] & -0.069 [-0.092, -0.045] \\
\midrule
Household costs & Left & 104 & -0.019 [-0.040, 0.002] & -0.075 [-0.094, -0.057] & -0.293 [-0.338, -0.249] \\
Household costs & Centre & 99 & -0.047 [-0.069, -0.025] & -0.072 [-0.094, -0.049] & -0.144 [-0.175, -0.113] \\
Household costs & Right & 97 & -0.094 [-0.121, -0.066] & -0.007 [-0.033, 0.020] & -0.002 [-0.037, 0.032] \\
\midrule
Immigration & Left & 104 & -0.004 [-0.022, 0.013] & -0.043 [-0.060, -0.026] & -0.165 [-0.203, -0.127] \\
Immigration & Centre & 95 & -0.063 [-0.087, -0.040] & -0.044 [-0.068, -0.020] & -0.152 [-0.183, -0.121] \\
Immigration & Right & 101 & -0.115 [-0.149, -0.082] & -0.004 [-0.033, 0.026] & -0.080 [-0.119, -0.041] \\
\midrule
Corruption & Left & 67 & -0.338 [-0.387, -0.290] & -0.329 [-0.376, -0.282] & -0.387 [-0.442, -0.331] \\
Corruption & Centre & 171 & -0.266 [-0.300, -0.232] & -0.243 [-0.272, -0.214] & -0.260 [-0.292, -0.228] \\
Corruption & Right & 62 & -0.326 [-0.372, -0.279] & -0.235 [-0.281, -0.188] & -0.164 [-0.205, -0.123] \\
\bottomrule\end{tabular}
\caption{Speaker-attribution effects on stance by topic and elicited ideological group, both frameworks. Cells report the mean effect [95\% CI].}
\label{tab:appendix-rq2-topic-groups}
\end{table*}

\begin{table*}[tbp]
\centering\small
\begin{tabular}{lccc}
\toprule
Outcome & Sánchez -- Feijóo & Sánchez -- Abascal & Feijóo -- Abascal \\
\midrule
Stance & 0.150 [0.128, 0.172] & 0.298 [0.264, 0.333] & 0.148 [0.124, 0.173] \\
Agreement & 0.288 [0.252, 0.324] & 0.559 [0.499, 0.619] & 0.271 [0.229, 0.314] \\
Legitimacy & 0.258 [0.226, 0.291] & 0.530 [0.473, 0.587] & 0.271 [0.231, 0.312] \\
Endorsement & 0.397 [0.357, 0.437] & 0.740 [0.672, 0.808] & 0.343 [0.293, 0.393] \\
Offensiveness & -0.022 [-0.036, -0.007] & -0.138 [-0.167, -0.109] & -0.117 [-0.143, -0.091] \\
\bottomrule\end{tabular}
\caption{Congruence contrasts by outcome, both frameworks and all topics. Cells report the estimate [95\% CI].}
\label{tab:appendix-rq2-outcomes}
\end{table*}

\subsection{Absolute response levels}
\label{app:levels}

The attribution effects are differences from the anonymous version of the same post. Table~\ref{tab:appendix-levels} gives the levels behind them. Without an author, half of the responses supported a post and a quarter rejected it (mean stance 0.24), so the penalties do not run into a floor or a ceiling. A name moved answers from support towards rejection, and the neutral share stayed near a quarter: under Abascal's name, support fell from 50\% to 40\% and rejection rose from 25\% to 34\%. Household-cost posts kept majority support under every name, Abascal's included (54\%). Corruption posts were close to neutral without an author ($-0.04$), and a majority rejected them under any name (54--57\%).

The two frameworks differ more in level than in effect. MFT personas were more favourable than IW personas in every condition (0.39 against 0.10 without an author), a gap larger than any attribution effect. The levels therefore describe how the personas of each framework answer under one model. They are not estimates of public support for the posts.

Table~\ref{tab:appendix-levels-profile} splits the anonymous level and the attribution effects by profile level. Within a topic, the levels follow their value profiles. Among MFT personas, the ``restriction and border control'' level supported the anonymous immigration posts (0.51) while the ``humanitarian inclusion'' level was slightly against them ($-0.07$), and the ``affordability and protection of basic needs'' level was the most favourable to household-cost posts (0.74). Abascal's penalty on the economy was largest at the MFT ``redistribution and social protection'' level ($-0.31$) and small at the ``merit and economic contribution'' level ($-0.04$). Levels cannot be compared across frameworks: on corruption, the high-accountability level was the most critical among IW personas ($-0.45$) and the most favourable among MFT personas (0.31).

Table~\ref{tab:appendix-levels-author} splits the posts by the leader who wrote them. Read anonymously, Sánchez's posts had a mean stance of 0.47, Feijóo's 0.28 and Abascal's $-0.35$. Abascal's posts had the lowest mean in every ideological group and under both frameworks, and even right personas were close to neutral on them ($-0.03$, against 0.40 for Feijóo's posts). The attribution effects compare every post with itself, so this difference in content does not enter them. It describes how the model received what each leader posts, not how the public did. Part of it may come from recognising the author, which is one reason why the anonymous condition is not fully anonymous (see Limitations).

\begin{table*}[tbp]
\centering\small
\setlength{\tabcolsep}{5pt}
\begin{tabular}{llrrrr}
\toprule
Personas & Topic & No author & Sánchez & Feijóo & Abascal \\
\midrule
Both & All topics & 0.24 (50/25/25) & 0.14 (46/23/31) & 0.14 (45/24/31) & 0.05 (40/26/34) \\
 & Economy and employment & 0.36 (54/27/19) & 0.34 (54/26/20) & 0.31 (52/27/21) & 0.17 (44/30/27) \\
 & Household costs & 0.47 (62/22/15) & 0.42 (60/21/19) & 0.42 (60/23/18) & 0.32 (54/24/22) \\
 & Immigration and border & 0.08 (43/23/35) & 0.02 (40/22/38) & 0.05 (41/23/36) & -0.05 (34/26/40) \\
 & Corruption & -0.04 (35/26/39) & -0.34 (23/20/57) & -0.30 (24/21/54) & -0.31 (24/22/55) \\
\midrule
IW & All topics & 0.10 (40/30/30) & 0.02 (38/27/36) & 0.02 (37/28/35) & -0.08 (31/29/40) \\
 & Economy and employment & 0.20 (43/33/24) & 0.17 (43/31/26) & 0.14 (40/33/26) & 0.01 (33/35/32) \\
 & Household costs & 0.34 (54/27/20) & 0.29 (52/25/23) & 0.28 (51/27/22) & 0.17 (44/28/27) \\
 & Immigration and border & 0.00 (37/26/37) & -0.06 (35/25/40) & -0.03 (35/26/39) & -0.17 (27/29/44) \\
 & Corruption & -0.21 (25/28/46) & -0.40 (19/22/59) & -0.38 (20/23/58) & -0.41 (18/22/59) \\
\midrule
MFT & All topics & 0.39 (59/21/20) & 0.27 (54/19/27) & 0.27 (53/20/26) & 0.19 (48/23/29) \\
 & Economy and employment & 0.52 (65/21/13) & 0.51 (65/21/14) & 0.48 (63/21/16) & 0.34 (55/25/21) \\
 & Household costs & 0.60 (71/18/11) & 0.55 (69/17/14) & 0.55 (69/18/13) & 0.47 (63/20/17) \\
 & Immigration and border & 0.16 (48/19/32) & 0.10 (46/18/36) & 0.13 (47/20/34) & 0.06 (41/23/35) \\
 & Corruption & 0.13 (45/23/32) & -0.27 (27/18/54) & -0.22 (29/20/51) & -0.21 (29/21/50) \\
\bottomrule
\end{tabular}
\caption{Absolute responses by condition: mean stance ($-1$ reject, $0$ neutral, $+1$ support) and, in parentheses, the percentage of support, neutral and reject answers, over the complete quartets. The attribution effects of RQ1 and RQ3 are the differences between a named column and the no-author column. 95\% confidence intervals of the means, two-way clustered by message and persona, have half-widths from 0.04 to 0.10.}
\label{tab:appendix-levels}
\end{table*}

\begin{table*}[tbp]
\centering\small
\setlength{\tabcolsep}{4pt}
\begin{tabular}{llrrrrrrrrrr}
\toprule
& & \multicolumn{5}{c}{IW personas} & \multicolumn{5}{c}{MFT personas} \\
\cmidrule(lr){3-7}\cmidrule(lr){8-12}
& & \multicolumn{2}{c}{No author} & \multicolumn{3}{c}{Effect of name} & \multicolumn{2}{c}{No author} & \multicolumn{3}{c}{Effect of name} \\
\cmidrule(lr){3-4}\cmidrule(lr){5-7}\cmidrule(lr){8-9}\cmidrule(lr){10-12}
Topic & Level & Mean & Supp.\ \% & S & F & A & Mean & Supp.\ \% & S & F & A \\
\midrule
Economy and employment & High & -0.18 & 20 & -0.11 & -0.08 & -0.13 & 0.51 & 65 & -0.06 & 0.00 & -0.04 \\
 & Mid & 0.36 & 53 & -0.02 & -0.02 & -0.16 & 0.54 & 66 & -0.02 & -0.04 & -0.19 \\
 & Low & 0.41 & 57 & 0.03 & -0.07 & -0.27 & 0.50 & 65 & 0.05 & -0.09 & -0.31 \\
\midrule
Household costs & High & 0.05 & 37 & -0.12 & -0.05 & -0.06 & 0.74 & 82 & -0.03 & -0.07 & -0.27 \\
 & Mid & 0.41 & 56 & -0.02 & -0.03 & -0.10 & 0.66 & 74 & -0.05 & -0.07 & -0.14 \\
 & Low & 0.55 & 67 & -0.01 & -0.09 & -0.35 & 0.40 & 57 & -0.08 & 0.00 & 0.02 \\
\midrule
Immigration and border & High & -0.11 & 34 & -0.01 & -0.05 & -0.21 & -0.07 & 36 & 0.00 & -0.04 & -0.13 \\
 & Mid & -0.01 & 35 & -0.07 & -0.03 & -0.14 & 0.04 & 41 & -0.05 & -0.04 & -0.12 \\
 & Low & 0.11 & 42 & -0.08 & -0.01 & -0.14 & 0.51 & 68 & -0.15 & -0.02 & -0.05 \\
\midrule
Corruption & High & -0.45 & 10 & -0.18 & -0.19 & -0.20 & 0.31 & 59 & -0.44 & -0.38 & -0.33 \\
 & Mid & -0.22 & 24 & -0.20 & -0.16 & -0.20 & 0.04 & 39 & -0.38 & -0.33 & -0.33 \\
 & Low & 0.04 & 42 & -0.20 & -0.16 & -0.21 & 0.02 & 37 & -0.36 & -0.34 & -0.34 \\
\bottomrule
\end{tabular}
\caption{Absolute responses and attribution effects on stance at each topic-conditioned profile level (Table~\ref{tab:appendix-profiles}), 50 personas per cell. No author: mean stance and percentage of support answers for the anonymous posts. Effect of name: Sánchez (S), Feijóo (F) or Abascal (A) minus no author. A level means a different value profile on each topic and in each framework, so levels are comparable only within a topic and framework. 95\% confidence intervals have half-widths from 0.06 to 0.13 for the means and from 0.01 to 0.06 for the effects.}
\label{tab:appendix-levels-profile}
\end{table*}

\begin{table*}[tbp]
\centering\small
\setlength{\tabcolsep}{5pt}
\begin{tabular}{llrrrrrrrr}
\toprule
& & & \multicolumn{4}{c}{Displayed condition} & \multicolumn{3}{c}{No author, by persona group} \\
\cmidrule(lr){4-7}\cmidrule(lr){8-10}
Personas & Written by & Posts & No author & Sánchez & Feijóo & Abascal & Left & Centre & Right \\
\midrule
Both & Sánchez & 504 & 0.47 & 0.45 & 0.41 & 0.27 & 0.67 & 0.38 & 0.35 \\
 & Feijóo & 360 & 0.28 & 0.12 & 0.15 & 0.08 & 0.34 & 0.15 & 0.40 \\
 & Abascal & 211 & -0.35 & -0.56 & -0.51 & -0.52 & -0.47 & -0.46 & -0.03 \\
\midrule
IW & Sánchez & 504 & 0.33 & 0.30 & 0.27 & 0.13 & 0.63 & 0.23 & 0.17 \\
 & Feijóo & 360 & 0.12 & -0.01 & 0.02 & -0.08 & 0.25 & 0.00 & 0.22 \\
 & Abascal & 211 & -0.49 & -0.61 & -0.58 & -0.62 & -0.59 & -0.57 & -0.25 \\
\midrule
MFT & Sánchez & 504 & 0.61 & 0.60 & 0.55 & 0.42 & 0.70 & 0.64 & 0.48 \\
 & Feijóo & 360 & 0.44 & 0.24 & 0.29 & 0.23 & 0.41 & 0.36 & 0.57 \\
 & Abascal & 211 & -0.22 & -0.50 & -0.44 & -0.42 & -0.38 & -0.33 & 0.20 \\
\bottomrule
\end{tabular}
\caption{Mean stance by the leader who actually wrote the post. Displayed condition: the anonymous post and the same post under each name. No author, by persona group: the anonymous post among personas who placed themselves on the left (1--4), centre (5--6) or right (7--10). 95\% confidence intervals have half-widths from 0.04 to 0.07 by condition and from 0.04 to 0.10 by group.}
\label{tab:appendix-levels-author}
\end{table*}

\FloatBarrier
\section{CIS Comparison Details}
\label{app:cis}
\begin{table*}[tbp]
\centering\small
\begin{tabular}{llcc}
\toprule
Leaders A -- B & Group & CIS (rating points) & Personas (effect on stance) \\
\midrule
Sánchez -- Feijóo & Left & 3.40 [3.37, 3.43] & 0.069 [0.055, 0.084] \\
Sánchez -- Feijóo & Centre & -1.06 [-1.10, -1.03] & -0.002 [-0.013, 0.009] \\
Sánchez -- Feijóo & Right & -3.50 [-3.54, -3.45] & -0.080 [-0.097, -0.064] \\
Sánchez -- Feijóo & Left $-$ right & 6.90 [6.84, 6.95] & 0.150 [0.128, 0.172] \\
\midrule
Sánchez -- Abascal & Left & 4.83 [4.80, 4.85] & 0.241 [0.212, 0.271] \\
Sánchez -- Abascal & Centre & 0.37 [0.34, 0.41] & 0.072 [0.053, 0.090] \\
Sánchez -- Abascal & Right & -3.11 [-3.16, -3.05] & -0.057 [-0.079, -0.036] \\
Sánchez -- Abascal & Left $-$ right & 7.93 [7.87, 8.00] & 0.298 [0.264, 0.333] \\
\midrule
Feijóo -- Abascal & Left & 1.43 [1.41, 1.44] & 0.172 [0.152, 0.192] \\
Feijóo -- Abascal & Centre & 1.44 [1.41, 1.47] & 0.074 [0.059, 0.089] \\
Feijóo -- Abascal & Right & 0.39 [0.34, 0.44] & 0.023 [0.007, 0.040] \\
Feijóo -- Abascal & Left $-$ right & 1.04 [0.99, 1.09] & 0.148 [0.124, 0.173] \\
\bottomrule
\end{tabular}
\caption{Differences between leaders within each group, and left minus right, among CIS respondents (difference in mean rating, pooled over 48 barometers) and personas ($\Delta_A-\Delta_B$ on stance, as in Table~\ref{tab:rq2-congruence}). Positive values mean that A is preferred to B. Cells report the estimate [95\% CI]. All twelve pairs have the same sign.}
\label{tab:appendix-rq4-pairs}
\end{table*}

\begin{table*}[tbp]
\centering\footnotesize
\setlength{\tabcolsep}{4pt}
\begin{tabular}{lcccccccccc}
\toprule
Placement & 1 & 2 & 3 & 4 & 5 & 6 & 7 & 8 & 9 & 10 \\
\midrule
Respondents & 20,817 & 12,167 & 26,512 & 19,255 & 40,743 & 17,161 & 15,636 & 11,343 & 2,632 & 9,498 \\
Personas & 1 & 63 & 129 & 65 & 239 & 47 & 40 & 152 & 41 & 14 \\
CIS order & S$>$F$>$A & S$>$F$>$A & S$>$F$>$A & S$>$F$>$A & F$>$S$>$A & F$>$A$>$S & F$>$A$>$S & F$>$A$>$S & A$>$F$>$S & A$>$F$>$S \\
Persona order & S$>$F$>$A & S$>$F$>$A & S$>$F$>$A & S$>$F$>$A & S$>$F$>$A & F$>$S$>$A & F$>$A$>$S & F$>$A$>$S & F$>$A$>$S & F$>$A$>$S \\
Same order & 48/48 & 48/48 & 48/48 & 48/48 & 0/48 & 9/48 & 46/48 & 38/48 & 26/48 & 15/48 \\
\bottomrule
\end{tabular}
\caption{Order of the leaders (S: Sánchez, F: Feijóo, A: Abascal) at each left--right placement among CIS respondents (pooled mean rating) and personas (speaker-attribution effect on stance), with the number of respondents and personas at each placement. The last row counts the barometers in which the CIS order is the same as the personas'. At placement 9 the pooled CIS ratings of Abascal and Feijóo differ by 0.02 points. The persona order at placement 1 rests on a single persona.}
\label{tab:appendix-rq4-placement}
\end{table*}

\begin{figure}[htbp]
\centering
\includegraphics[width=\columnwidth]{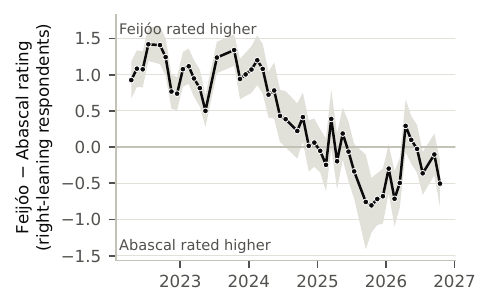}
\caption{Difference between the mean ratings of Feijóo and Abascal among right CIS respondents (7--10), barometer by barometer, with 95\% confidence intervals. The order among right personas (Feijóo above Abascal) matches the barometers above zero.}
\label{fig:appendix-rq4-trend}
\end{figure}

The 48 barometers are every monthly CIS barometer from April 2022 to October 2026; the CIS published none in August, nor in June and September 2023, around the general election. The leader item is \textit{¿Qué valoración le merece cada uno/a de ellos/as? Puntúelos/as de 1 a 10, sabiendo que el 1 significa que lo/a valora `muy mal' y el 10 que lo/a valora `muy bien'}, asked about each leader the respondent knows; it is the only leader item that covers all three leaders in every barometer, as the confidence items exclude Abascal. The ideology item is the self-placement question the personas answered (Appendix~\ref{app:ideology}). Of 193,263 respondents, 8,599 did not place themselves on the scale and 8,900 did not rate all three leaders (because they did not know one, or answered ``don't know'' or did not answer); the remaining 175,764 were analysed, so the leader means and the differences between leaders come from the same respondents. Ratings were weighted with each barometer's national weight, which the CIS adjusted for education from February 2024. Standard errors are linearised for a weighted mean within each barometer, and the barometers are treated as independent samples when pooled. The intervals ignore the clustering of the sample design and are, therefore, somewhat narrow; the comparison uses only the order and sign of the estimates.

Table~\ref{tab:appendix-rq4-pairs} gives the twelve differences between leaders among respondents and personas, Table~\ref{tab:appendix-rq4-placement} the order of the leaders at each of the ten placements, and Figure~\ref{fig:appendix-rq4-trend} the Feijóo-minus-Abascal difference among right respondents barometer by barometer. Its confidence interval lies entirely below zero in 8 barometers from September 2025 onwards.

\end{document}